\documentclass[10pt,twocolumn,letterpaper]{article}

\usepackage[utf8]{inputenc}

\usepackage{iccv}
\usepackage{times}
\usepackage{epsfig}
\usepackage{graphicx}
\usepackage{amsmath}
\usepackage{amssymb}

\usepackage{booktabs}
\usepackage{caption}
\usepackage{subcaption}
\usepackage{xcolor}
\usepackage{xfrac}
\usepackage{pifont}
\usepackage{multirow}
\usepackage[numbers,sort,compress]{natbib}
\usepackage{flushend}
\usepackage[subtle]{savetrees}
\newcommand{\R}{\mathbb{R}}
\usepackage{dsfont}
\newcommand\norm[1]{\left\lVert#1\right\rVert}
\usepackage{float}

\usepackage{setspace}

\usepackage{iccv}
\usepackage{times}
\usepackage{epsfig}
\usepackage{graphicx}
\usepackage{amsmath}
\usepackage{amssymb}

\usepackage[pagebackref=true,breaklinks=true,letterpaper=true,colorlinks,bookmarks=false]{hyperref}

\def\iccvPaperID{8001} 

\iccvfinalcopy 

\begin{document}


\title{Bayesian Triplet Loss: Uncertainty Quantification in Image Retrieval}

\author{Frederik Warburg\textsuperscript{$\dagger$}, Martin Jørgensen\textsuperscript{$\ddagger$}, Javier Civera\textsuperscript{$\S$}, and Søren Hauberg\textsuperscript{$\dagger$} \\
\textsuperscript{$\dagger$}Technical University of Denmark, \textsuperscript{$\ddagger$}University of Oxford,  \textsuperscript{$\S$}University of Zaragoza \\
\normalsize {\fontfamily{pcr}\selectfont
\textsuperscript{$\dagger$}\{frwa,sohau\}@dtu.dk, 
\textsuperscript{$\ddagger$}martinj@robots.ox.ac.uk, \textsuperscript{$\S$}jcivera@unizar.es}
}

\maketitle

\begin{abstract}

Uncertainty quantification in image retrieval is crucial for downstream decisions, yet it remains a challenging and largely unexplored problem. Current methods for estimating uncertainties are poorly calibrated, computationally expensive, or based on heuristics.
We present a new method that views image embeddings as stochastic features rather than deterministic features.
Our two main contributions are (1) a likelihood that matches the triplet constraint and that evaluates the probability of an anchor being closer to a positive than a negative; and (2) a prior over the feature space that justifies the conventional $l_2$ normalization.
To ensure computational efficiency, we derive a variational approximation of the posterior, called the Bayesian triplet loss, that produces state-of-the-art uncertainty estimates and matches the predictive performance of current state-of-the-art methods. 

\end{abstract}

\section{Introduction}

Image-based retrieval systems show impressive performance in challenging tasks such as face verification~\cite{DBLP:journals/corr/SchroffKP15,DBLP:journals/corr/abs-1804-06655,6909616}, instance retrieval~\cite{zheng2017sift}, landmark retrieval~\cite{DBLP:journals/corr/abs-1906-04087} and place recognition~\cite{DBLP:journals/corr/abs-1711-02512, DBLP:journals/corr/ArandjelovicGTP15}. 
These systems typically embed images into high-level features and retrieve with a nearest neighbor search. 
While this is efficient, the retrieval comes with no notion of confidence, which is particularly problematic in safety-critical applications.
For instance, a self-driving car relying on visual place recognition should be able to filter out place estimates drawn from uninformative images. In a less critical but still relevant application, quantifying the retrieval uncertainty can significantly improve the user experience in human-computer interfaces by not showing low-confidence results for a query.

\begin{figure}
    \centering
    \includegraphics[width=0.85\linewidth]{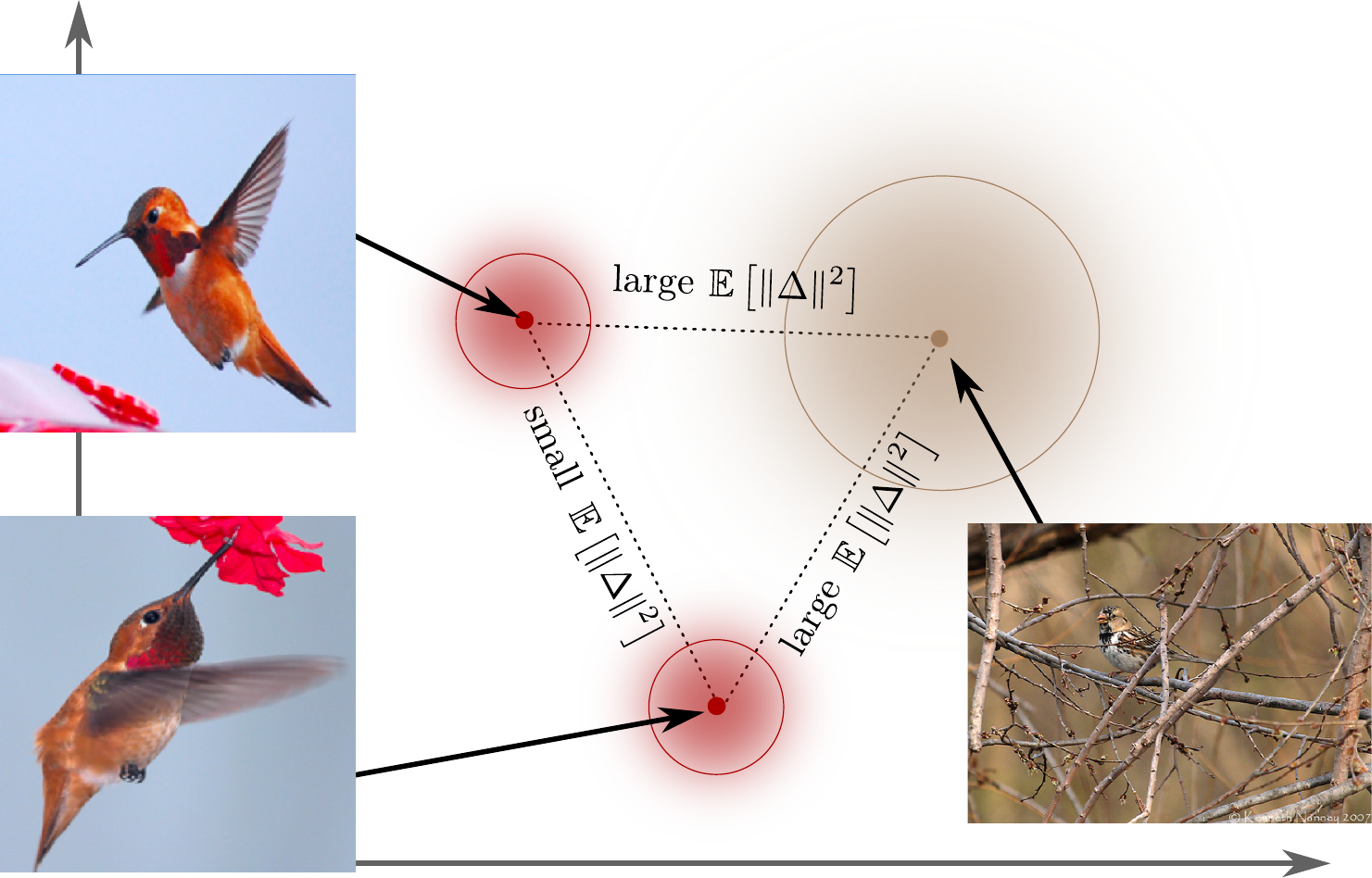}
    \caption{We model embeddings as distributions rather than point estimates, such that data uncertainty is propagated to retrieval. We phrase a Bayesian model that mirrors the triplet loss, which enables us to learn the stochastic features. 
    }
    \label{fig:teaserfigure}
\end{figure}

Practical retrieval systems do not have a small set of predefined classes as output targets, but rather need high-level features that generalize to unseen classes. For instance, a visual place recognition system may be deployed in a city in which it has not been trained~\cite{Warburg_CVPR_2020}. This is achieved by keeping the encoder fixed and relying on nearest neighbor searches. This pipeline does not easily match current methods for posterior inference, and current uncertainty estimators for retrieval are often impractical and heuristic in nature. To construct a fully Bayesian retrieval system that fits with existing computational pipelines, we first recall the elementary equation
\begin{align}
  \mathbb{E}\left[ \| \Delta \|^2 \right] = \left\| \mathbb{E}\left[\Delta\right] \right\|^2 + \mathrm{trace}\left( \mathrm{cov}[\Delta] \right), \qquad \Delta \in \mathbb{R}^D,
\end{align}
which follows directly from the definition of variance. From this we see that the expected squared distance between two random features, $\mathbb{E}\left[ \| \Delta \|^2 \right]$, grows with the covariance of such distance, $\mathrm{cov}[\Delta]$, which in turn depends on the uncertainty of the features (Figure \ref{fig:teaserfigure}). This intuition forms the basis of this paper.

\textbf{In this paper} we propose to use stochastic image embeddings instead of the usual deterministic ones. Given an image $X$, we consider the posterior distribution over possible features $P(F|X)$. From this distribution we get direct uncertainty estimates and can assign probabilities to events such as `two images belonging to the same place'. To realize this, we derive a likelihood corresponding to the probability that the conventional triplet constraint is satisfied, and a prior over the feature space that mimics conventional $l_2$ normalization. To build a system that is computationally efficient at both train and test time, we derive a variational approximation to the posterior $P(F|X)$, such that in practice, we encode an image to a distribution in feature space.
Across several datasets, we show that the proposed model matches the state-of-the-art in predictive performance, while attaining state-of-the-art uncertainty estimates.

\section{Related Work}

\textbf{Image retrieval} has been a popular research problem in the last years due to its many applications \cite{smeulders2000content,zheng2017sift}.
Early approaches relied on handcrafted local features aggregated mainly by bag-of-words \cite{sivic2003video,philbin2007object,jegou2010aggregating}. More recent models are composed by a deep convolutional backbone followed by an aggregation layer~\cite{DBLP:journals/corr/ArandjelovicGTP15,DBLP:journals/corr/abs-1711-02512} that map images to low dimensional embeddings based on their content similarity \cite{DBLP:journals/corr/SchroffKP15,DBLP:journals/corr/abs-1804-06655,6909616,DBLP:journals/corr/abs-1906-04087,DBLP:journals/corr/abs-1711-02512, DBLP:journals/corr/ArandjelovicGTP15,noh2017large,wang2018cosface,revaud2019learning,brown2020smooth}. The most similar images to a query are found by (approximate) nearest neighbor methods.



Retrieval systems apply either classification losses (e.g., \cite{zhai2018classification,liu2017sphereface}) or metric losses (e.g., \cite{gordo2017end}). Metric losses operate on the relationships between samples in a batch, while classification losses include a weight matrix that transforms the embeddings into vectors of class logits~\cite{musgrave2020metric}. Our focus is on metric losses, of which the contrastive loss~\cite{contrastiveloss} is the most fundamental one. However, this loss has the limitation that the same margin thresholds apply to all training pairs, even though there may be a large variation in their similarities. The triplet loss~\cite{tripletloss} accounts for such varying interclass dissimilarities by solely constraining an anchor $a$ to be closer to a positive $p$ than a negative $n$ minus a margin $m$, 
\begin{equation}
    \norm{ a - p}^2 < \norm{ a - n}^2 - m.
    \label{eq:tripletloss}
\end{equation}


Many works have extended the contrastive and triplet losses to incorporate more structural information about the embedding space, for example the quadruplet loss~\cite{chen2017beyond}, N-pair loss~\cite{sohn2016improved}, angular loss~\cite{wang2017deep}, margin loss~\cite{wu2017sampling}, signal-to-noise ratio (SNR) contrastive loss~\cite{yuan2019signal} and multi-similarity (MS) loss~\cite{wang2019multi}. These methods are often supported by heuristics or empirical experiments, but lack a theoretical grounding. Additionally, in a recent study, \citet{musgrave2020metric} show that these more complex loss functions offer only marginal improvements over the contrastive and triplet losses. For this reason we focus on the triplet loss, but in principle our approach can be extended to mirror other losses.

\textbf{Uncertainty in deep networks} is hard to quantify due to the large number of parameters~\cite{gal2016uncertainty}. Uncertainty quantification is currently being studied in the context of many computer vision tasks, among others depth completion~\cite{eldesokey2020uncertainty,gustafsson2020evaluating}, semantic segmentation~\cite{kendall2015bayesian,kohl2018probabilistic,gustafsson2020evaluating}, object detection~\cite{choi2019gaussian,gustafsson2020energy}, object pose estimation~\cite{okorn2020learning} and multi-task learning~\cite{kendall2018multi}. In practice, Bayesian approximations such as deep ensembles~\cite{lakshminarayanan2017simple}, Monte Carlo dropout (MC dropout)~\cite{gal2016dropout} and conditional autoencoders~\cite{kingma2013auto} have shown most promise. Although scalable~\cite{gustafsson2020evaluating}, these methods do not directly apply to image retrieval, as models typically do not have proper likelihood functions. 

\citet{der2009aleatory} identify the sources of predictive uncertainties as the model (epistemic uncertainty) and the data (aleatoric uncertainty). The latter can be divided into homoscedastic (constant for all input data) and heteroscedastic (variable depending on the particular input). We focus on heteroscedastic uncertainty as this is especially relevant for image retrieval as illustrated in Figure~\ref{fig:teaserfigure}.

\textbf{Uncertainty quantification for image retrieval} using deep networks is a challenging and under-addressed topic. Learning stochastic embeddings rather than deterministic ones has been addressed \ie for images~\cite{DBLP:journals/corr/abs-1810-00319, DBLP:journals/corr/abs-1904-09658, chang2020data}, human poses~\cite{DBLP:journals/corr/abs-1912-01001} and cross-modal data~\cite{DBLP:journals/corr/abs-1906-04402, chun2021probabilistic}. Most prior work has focused on classification~\cite{chang2020data} or pairwise losses. \citet{DBLP:journals/corr/abs-1810-00319} use Monte Carlo (MC) sampling from learned Gaussian distributions to evaluate a matching probability between a pair of images. Their approach is successful for low dimension embeddings ($D = 3$). The expensive Monte Carlo sampling can be avoided by directly optimizing the likelihood that two Gaussian embeddings belong to the same class~\cite{DBLP:journals/corr/abs-1904-09658}. In our work, we extend this intuition to triplets.
Triplet losses, contrary to pairwise ones, are known to address varying interclass similarities and dissimilarities \cite{musgrave2020metric}. 
\citet{DBLP:journals/corr/abs-1912-01001} suggest a ratio of two binary cross entropy terms. Taking the log of these gives an expression on the form of the triplet loss. However, their approach relies on MC samples. \citet{DBLP:journals/corr/abs-1901-07702} cast the triplet loss as a regression loss and estimate epistemic uncertainty with MC dropout. In later work, \citet{DBLP:journals/corr/abs-1902-02586} propose to learn the heteroscedastic uncertainty using a noise parameter per image. In contrast to our proposal, no model is provided under which the proposed triplet losses is a proper likelihood. We benchmark our method against a MC sampling method (triplet regression with MC dropout~\cite{DBLP:journals/corr/abs-1901-07702}) and a implicit method (triplet regression that implicitly learns a noise parameter~\cite{DBLP:journals/corr/abs-1902-02586}), showing significantly better uncertainty estimates in several datasets while matching the predictive performance of the standard triplet loss. 


\section{Bayesian Triplet Loss}

We propose to embed images as distributions rather than point estimates. Given these stochastic embeddings, we ask: What is the probability that an anchor is closer to a positive than a negative, \ie
\begin{equation}
\label{eq:probtriplet1}
    P\left(\|a - p\|^2 < \|a - n\|^2 - m\right),
\end{equation}
which is the probabilistic equivalent of the triplet constraint~\eqref{eq:tripletloss}. To realize this idea, we first derive a likelihood function corresponding to Eq.~\ref{eq:probtriplet1}. 
As we want this to follow the intuition of the triplet loss closely, our likelihood operates on triplets of images. We define our likelihood for all triplets in the dataset,
\begin{equation}
    \mathcal{L}(\Omega)=\prod_{X \in \Omega} \prod_{Y \in \Omega} \prod_{Z \in \Omega} P(I(X, Y, Z) | X, Y, Z),
\end{equation}
where $X$, $Y$, $Z$ are images from the dataset and $I$ is the label of a triplet (referred to as triplet label). We note that the triplet labels can take three values with respectively $1$, $2$ or $3$ images from the same class. Since $I(X,Y,Z)$ only takes values on a discrete finite set, we define our \emph{likelihood function} as the multinomial distribution. 
%
The traditional triplet loss ignores the situations where all images are from the same or different classes, as these situations are not informative. Making a similar modeling choice, the likelihood reduces to only consider triplets with one pair 
\begin{align}\label{eq:likelihood}
\begin{split}
    P(I(X,Y,Z)| X, Y, Z) =  \mathcal{P}^{\mathds{1}\{I(X,Y,Z) \, = \, 2\}}.
\end{split}
\end{align}

Thus, all probability mass is in triplets where two images are from the same class and one from a different class, just as with the traditional triplet loss. Using the standard triplet notation, we set $\mathcal{P}$ equal to Eq. \ref{eq:probtriplet1}, thus we derived a proper likelihood function that describes the probability of an anchor being closer to a positive than a negative.
We adopt this simpler notation throughout the remainder of the paper to make it clear which images are from the same class, but emphasize that the likelihood is defined for \emph{all} triplets via Eq.~\ref{eq:likelihood}.

\begin{figure}[ht]
    \centering
    \includegraphics[width=\hsize]{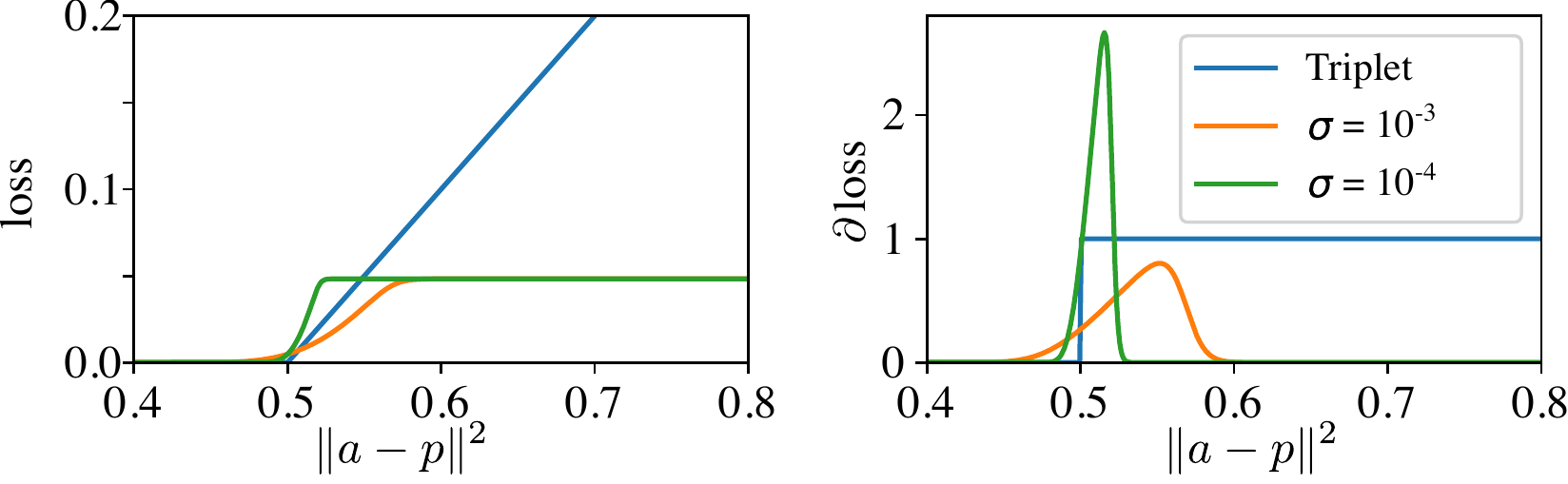}
    \caption{The traditional triplet loss (blue) compared to our negative log likelihood (orange and green). The negative log likelihood is smooth and bounded, yielding better robustness.}
    \label{fig:losss}
\end{figure}

Figure~\ref{fig:losss} shows the proposed negative log-likelihood compared to the traditional triplet loss. Our likelihood is smooth and bounded, making it more robust to outliers. We experience similar training time as with the traditional triplet loss and have not experienced zero-gradients that block learning.

\begin{figure}[]
    \centering
    \includegraphics[width=\linewidth]{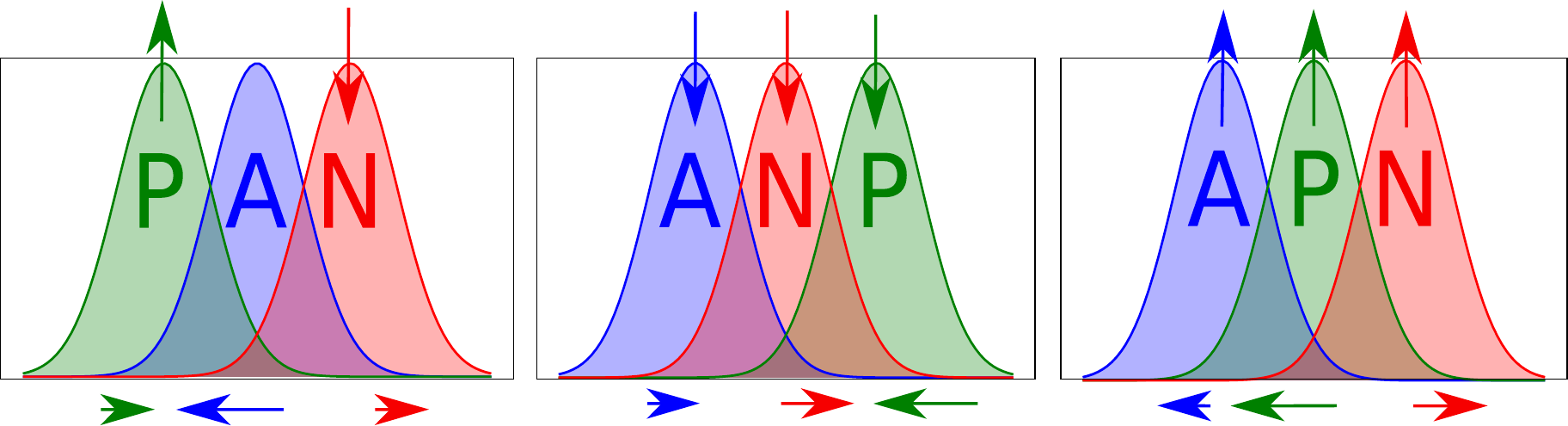}
    \caption{Intuition for the Bayesian triplet loss in three 1D scenarios. The arrows below the figures indicate the gradient direction and magnitude of the means, while the arrows above the distributions indicate the gradients of the variances (downwards indicate more spread, upwards means more peaky).}
    \vspace{-2mm}
    \label{fig:intuition}
\end{figure}

\subsection{The Triplet Likelihood}
Having the likelihood form in place, we proceed to arrive at explicit expressions.
We assume that the embeddings are isotropic Gaussians rather than points, such that $x~\sim~\mathcal{N}(\mu_x, \; \sigma_x^2 I)$, where $x \in \{a,\; p,\; n\}$ (see Figure~\ref{fig:teaserfigure}). This will be justified in the next section. Rearranging Eq.~\ref{eq:probtriplet1} gives
\begin{align}
\label{eq:probtriplet}
    P(\|a - p\|^2 &- \|a - n\|^2 < -m) = P(\tau < -m), \\
    \text{where } \tau &= \sum_{d=1}^D (a_d - p_d)^2  - (a_d - n_d)^2, \label{eq:tau}
\end{align}
and $D$ is the feature dimension. The squared distance between two normally distributed random variables follows a scaled non-central $\chi^2$-distribution \cite{muirhead2009aspects}.
The likelihood \eqref{eq:probtriplet} is a linear combination of two such distributed squared distances, which does not have a known density, and we resort to approximations.

By the central limit theorem \cite{montgomery2007applied}, $\tau$ will approximate a Gaussian distribution for large $D$, \ie
\begin{equation}\label{eq:central-limit}
    \lim_{D \rightarrow \infty} P\left(\frac{\tau - \mu}{\sigma} < -m \right) = \Phi(-m),
\end{equation}
where $\Phi$ is the CDF of the standard normal distribution, and $\mu$ and $\sigma$ are the mean and standard deviation of $\tau$.
In the supplements, we experimentally demonstrate that this approximation is remarkably accurate even in low dimensions.

We still need to find the leading two moments of $\tau$ to apply this approximation. We provide detailed derivations in the supplements, and here only sketch the steps.

\textbf{The mean} is determined as 
\begin{equation}
  \mathbb{E}[\tau]
    = \mathbb{E}[p(p - 2a)] - \mathbb{E}[n(n - 2a)].
\end{equation}
We exploit the symmetry and write first
\begin{equation}
  \mathbb{E}[p(p - 2a)]
    = \mathbb{E}[p^2] - 2\mathbb{E}[ap]
     = \mathbb{E}[p^2] - 2\mathbb{E}[a]\mathbb{E}[p],
\end{equation}
since $a$ and $p$ are independent. 
Using identical arguments for the second term of $\mathbb{E}[\tau]$ we get
\begin{align}
\begin{split}
  \mathbb{E}[\tau]
    &= \mathbb{E}[p^2] - 2\mathbb{E}[a]\mathbb{E}[p]
     - \mathbb{E}[n^2] + 2\mathbb{E}[a]\mathbb{E}[n] \\
    &= \mathbb{E}[p^2] - \mathbb{E}[n^2]
     - 2\mathbb{E}[a](\mathbb{E}[p] - \mathbb{E}[n]) \\
    &= \mu_p^2 + \sigma_p^2 - \mu_n^2 - \sigma_n^2
     - 2\mu_a(\mu_p - \mu_n).
\end{split}
\end{align}

\textbf{The variance} requires a longer derivation, so we only present the result here,
\begin{align}
\frac{\text{Var}(\tau)}{2} &= \sigma^2_p(\sigma^2_p + 2\mu^2_p) + \sigma_n^2(\sigma_n^2 + 2\mu^2_n) - 4\sigma^2_a\mu_p\mu_n \nonumber\\
&- 2\mu_a\left(\mu_a(\mu_p^2+\mu_n^2) - 2\mu_p\sigma^2_p - 2\mu_n\sigma^2_n\right) \\&+ 2(\sigma^2_a+\mu^2_a)\left( (\sigma_p^2 + \mu^2_p) + (\sigma^2_n + \mu^2_n)\right). \nonumber
\end{align}
Thus, given a mean and a variance estimate for each image in the triplet, we can analytically compute the mean and variance of $\tau$.
Then the likelihood \eqref{eq:likelihood} is evaluated by a Gaussian likelihood with these parameters. 

\textbf{The intuition} of the likelihood function is shown in Figure~\ref{fig:intuition} for three 1D scenarios. In the left figure, the ordering is correct (anchor is closer to the positive than the negative). The gradients w.r.t the variances are negative, thus reducing the uncertainty of each stochastic embedding (indicated with the arrows above the distributions). In the center figure, the ordering is incorrect (anchor is closer to the negative than the positive) resulting in higher uncertainties. The arrows below the plots indicate the gradient direction and magnitude w.r.t.\ the means. In all scenarios the mean of the anchor and positive are attracted, while the mean of the negative is repelled. 

\subsection{Normalization Priors}
It is common practice in image retrieval to $l_2$-normalize the embeddings as this often boosts retrieval performance \cite{DBLP:journals/corr/ArandjelovicGTP15,DBLP:journals/corr/abs-1711-02512,Radenovic-TR-2019-01}. A further practical benefit is that for $l_2$-normalized vectors, the Euclidean distance and the cosine similarity have a monotonic relation, and thus can be interchanged without altering the retrieval order. The cosine similarity is computationally efficient as it reduces to the dot product for normalized vectors~\cite{Radenovic-TR-2019-01}. 
We investigate two priors to imitate this normalization.

In high dimensions, the standard Gaussian distribution concentrates around a sphere of radius $D$. Hence, we can mimic the $l_2$ normalization by imposing a Gaussian prior over the embeddings. In particular, the prior $p(x) = \mathcal{N}(x|0,\tfrac{1}{D}I)$ concentrates around the unit sphere, and can therefore be seen as an \textit{implicit} $l_2$ normalization.
We also consider an \textit{explicit} normalization prior, by choosing a uniform prior over the unit sphere.


\subsection{The Approximate Posterior}
The posterior embedding is generally intractable, and we resort to variational approximations for computational efficiency~\cite{blei2017variational}.
We choose a parametrized approximate posterior $q$ as an isotropic distribution from the same family as the prior.
With the Gaussian prior we choose $q(X) =\mathcal{N}(\mu_X,\sigma^2_X I)$. With the uniform spherical prior we choose the approximate posterior as a von Mises Fisher distribution $q(X) = \text{vMF}(\mu_X, \kappa_X)$ \cite{dhillon2003modeling}. The distribution parameters are here described by the neural network.

In the supplements we derive the Expected Lower Bound (ELBO) for the marginal likelihood to be the right-hand side here
\begin{align}\label{eq:the-elbo}
    \log P( I(X, Y, Z) )  &\geq
    \mathbb{E}_{q(X)q(Y)q(Z)}[\log P(I(X, Y, Z) | X, Y, Z) ]  \nonumber\\
    & -\text{KL}(q(X) \| p(X)) - \text{KL}(q(Y) \| p(Y)) \nonumber\\ 
    & -\text{KL}(q(Z) \| p(Z)).
\end{align}
With the chosen distribution families, the KL divergences have closed-form expressions~\cite{diethe2015note, murphy2012machine}.


\section{Network Architecture and Training}

\begin{figure}[ht]
    \centering
    \includegraphics[width=\hsize]{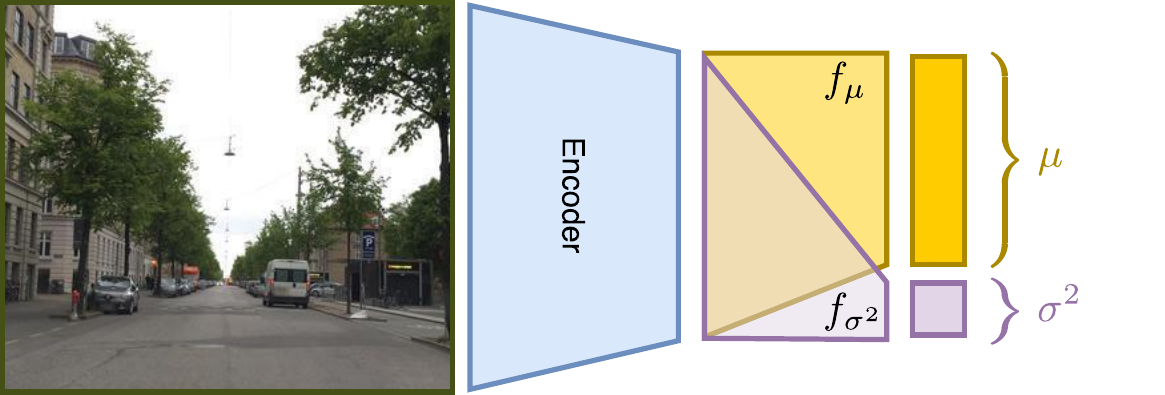}
    \caption{Overview of our network architecture.}
    \vspace{-6mm}
    \label{fig:encoder}
\end{figure}

For each image we learn an isotropic distribution rather than a point embedding. We treat both Gaussian and von Mises Fisher embeddings identically, and here only describe the Gaussian setup. Similar to \citet{DBLP:journals/corr/abs-1902-02586}, we use a shared backbone network followed by a mean and a variance head (see Fig.~\ref{fig:encoder}). The mean head is a generalized mean (GeM)~\cite{DBLP:journals/corr/abs-1711-02512} aggregation layer followed by a fully connected layer that outputs $\mu \in \R^D$. The variance head consists of a GeM layer followed by two fully connected layers with a ReLU activation function. We found it was advantageous to estimate $\sigma^2$ with a softplus activation rather than estimating $\log \sigma^2$. We have separate GeM layers for the variance and mean heads, as we found it beneficial to learn different $p$-norms for the variance head and mean head. 

In real world applications this trade-off between predictive performance and uncertainty quantification is important. Therefore, we ensure that the number of output parameters is the same for probabilistic and non-probabilistic models, such that $D_{\mu}\!+\!D_{\sigma}\!=\!D$. We focus on isotropic distributions and set $D_{\sigma}\!=\!1$. 
For the triplet loss, we follow common practice and $l_2$-normalize the point estimates, that is $x / \|x\|_2 \in \R^{D}$. For the Bayesian triplet loss, we $l_2$-normalize the mean embedding $\mu / \|\mu\|_2 \in \R^{D_{\mu}}$ for the uniform prior, and scale with a single positive trainable parameter for the Gaussian prior. 



We use a hard negative mining strategy similar to \citet{DBLP:journals/corr/ArandjelovicGTP15}. Given a query image, we find the closest negative images in a cache. We only present the model with the triplets that violate the triplet constraint \eqref{eq:tripletloss}. We update the cache with $5000$ new images every $1000$ iterations.  \citet{DBLP:journals/corr/ArandjelovicGTP15} and \citet{Warburg_CVPR_2020} report the importance of updating the cache regularly to avoid overfitting. This speeds up learning by reducing the number of trivial examples presented to the model.\looseness=-1

\begin{figure*}[htp!]
     \centering
    \includegraphics[width=0.9\textwidth]{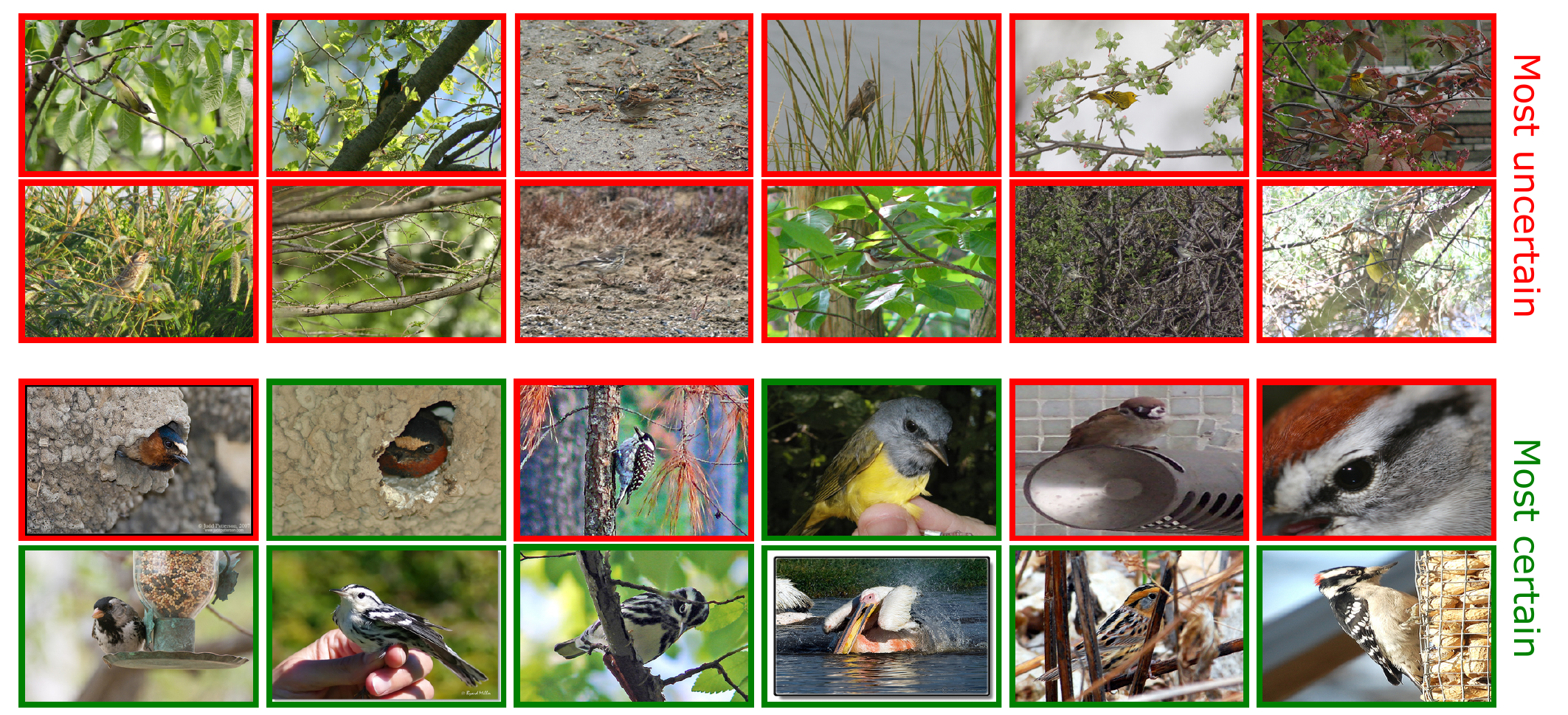}
     \caption{Query images for which our Bayesian embedding gives the highest (two first rows) and lowest uncertainty (two last rows). Scenes associated with high uncertainty mostly correspond to scenes where birds blend in with the background and are hardly discernible. The two most certain ones correspond to Cliff Swallows, easily discernible by the characteristic mud nests that they cement to walls or cliffs. In all images, birds stand out from the background and have unique patterns.}
     \vspace{-4mm}
     \label{fig:most_certain_bird}
\end{figure*}

\section{Evaluation Metrics}
The \textbf{Recall at k} (R@k) measures the number of queries that have at least one positive among their closest $k$ neighbors. This is a commonly used metric for image retrieval. However, this metric does not take into account the ratio of positives and negatives among the neighbors. Therefore, we also evaluate the \textbf{Mean Average Precision at k} (mAP@k) which measures the precision of the $k$ closest neighbors~\cite{DBLP:journals/corr/ArandjelovicGTP15}. These metrics evaluate the predictive performance of our models. 

The \textbf{Expected Calibration Error} (ECE) describes how well a model's uncertainties correspond with its predicted accuracy, and is a common metric to measure model calibration in classification tasks~\cite{DBLP:journals/corr/GuoPSW17,gustafsson2020evaluating}. The predictions are divided into $M$ equally spaced bins based on their confidences. For each bin $B_{m}$, the accuracy is compared to the model confidence, and weighted by the bin size. We reformulate this metric to fit for retrieval problems. Confident queries should have a high mAP and unconfident queries low mAP. We can therefore let ECE@k measure the weighted distance between mAP@k and the $M$\textsuperscript{th} percentiles of the variance. We set $M=10$. 
%
\begin{equation}
\mathrm{ECE}=\sum_{m=1}^{M} \frac{\left|B_{m}\right|}{n}\left|\operatorname{mAP@k}\left(B_{m}\right)-\operatorname{conf}\left(B_{m}\right)\right|.
\end{equation}    


\section{Experiments and Results}

We conduct experiments across three challenging image retrieval datasets. For each of the experiments, we compare the traditional triplet loss~\cite{tripletloss} with the proposed Bayesian triplet loss with the Gaussian prior (Bayes Triplet) and with the von-Mises Fisher prior (Bayes vMF). We also compare our model's uncertainty estimates with those produced by the triplet regression~\cite{DBLP:journals/corr/abs-1902-02586} and MC dropout~\cite{DBLP:journals/corr/abs-1901-07702}. We evaluate on two strong backbones, namely Resnet50~\cite{DBLP:journals/corr/HeZRS15} ($D = 2048$) and the large Densenet161~\cite{DBLP:journals/corr/HuangLW16a} ($D = 2208$), to illustrate that the Bayesian triplet loss provides calibrated uncertainty estimates across different architectures. Densenet161 is chosen specifically as it applies dropout in the backbone, which allows us to compare the uncertainty estimates with MC dropout. For comparison with triplet regression~\cite{DBLP:journals/corr/abs-1901-07702} we consistently use a dropout rate of $0.2$ as reported in ~\cite{DBLP:journals/corr/abs-1901-07702}. All models are implemented in Pytorch and trained with the Adam optimizer \cite{kingma2017adam} with learning rate $10^{-5}$, weight decay $0.001$ and an exponential learning rate scheduler decreasing the learning rate by $1 \%$ per epoch. We use batches of $25$ triplets, each triplet consisting of one anchor, one positive and five negative images. We resize the images to $224\times224$ in all the experiments. During training, we augment the data with random rotations (up to $10^\circ$), resized cropping ($[0.4; 1]$ of image size), color jitter and horizontal flipping. We use a KL-scale factor $10^{-6}$ in all experiments.

\begin{figure*}[ht]
     \centering
     \includegraphics[width=0.9\hsize]{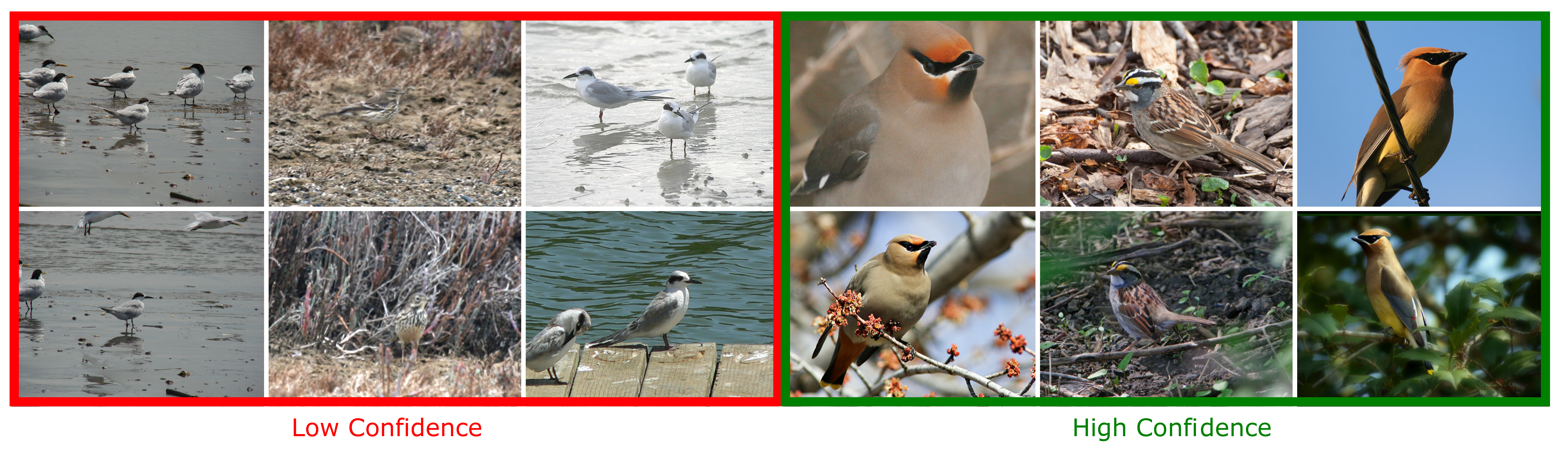}
     \vspace{-4mm}
     \caption{Six query images (top) and their NN (bottom), all true positives. Our Bayesian model assigns low confidence to images with multiple birds or where birds blend in with their surroundings. Birds with distinguishable patterns or colors have high confidence.}
     
     \label{fig:bird_confidence}
\end{figure*}

\begin{table*}[t]
\begin{center}
\resizebox{!}{0.12\textwidth}{%
\begin{tabular}{ll|ccccccccc}
                         & & R@1  & R@5  & R@10 & M@1  & M@5  & M@10 &ECE@1			&ECE@5			&ECE@10\\ \hline \hline
\multirow{6}{*}{\rotatebox{90}{ResNet50}} 
&Triplet			&\bf0.648			&\bf0.864			&\bf0.917			&\bf0.648			&\bf0.505			&\bf0.440			&			&			&  \\
&TripReg			&0.643			&0.863			&0.916			&0.643			&0.503			&0.437			&0.196			&0.331			&0.397 \\
&Bayes vMF		&0.635			&0.855			&0.911			&0.635			&0.492			&0.426			&0.138			&0.064			&0.089 \\
&Bayes Triplet		&0.612			&0.842			&0.902			&0.612			&0.467			&0.397			&\bf0.119			&0.037			&0.099 \\
&Bayes Triplet (Exp)		&0.632			&0.857			&0.912			&0.630			&0.489			&0.424			&0.137			& \bf0.020			& \bf0.072 \\ \hline
\multirow{6}{*}{\rotatebox{90}{Densenet161}} 
&Triplet			&\bf0.717			&0.894			&0.935			&\bf0.717			&\bf0.598			&\bf0.537			&			&			& \\
&Triplet (MC=50)			&0.349			&0.591			&0.700			&0.349			&0.200			&0.143			&0.290			&0.428			&0.480 \\
&TripReg	 				&0.711			&0.897			&\bf0.939			&0.711			&0.587			&0.524			&0.181			&0.302			&0.363 \\
& Bayes vMF & 0.713&	\bf0.898&	0.938&	0.713&	0.590&	0.528&	\bf0.022&	0.139&	0.200 \\
&Bayes Triplet		&0.683			&0.879			&0.926			&0.683			&0.564			&0.502			&0.176			& 0.059			&\bf0.023 \\
& Bayes Triplet (Exp)		&0.617			&0.838			&0.902			&0.617			&0.494			&0.437			& 0.101			&\bf0.025			&0.075 \\
\end{tabular}
}
\caption{Recall (R), Mean Average Precision (M) and Expected Calibration Error (ECE) at $1$, $5$ and $10$ on the CUB200 dataset. Bayes Triplet (Exp) refers to the Expected distance rather than the Euclidean distance was used for nearest neighbor search.}
\vspace{-4mm}
\label{tab:retrieval_performance_cub}
\end{center}
\end{table*}

\begin{figure}[t]
    \centering
    \includegraphics[width=0.9\hsize]{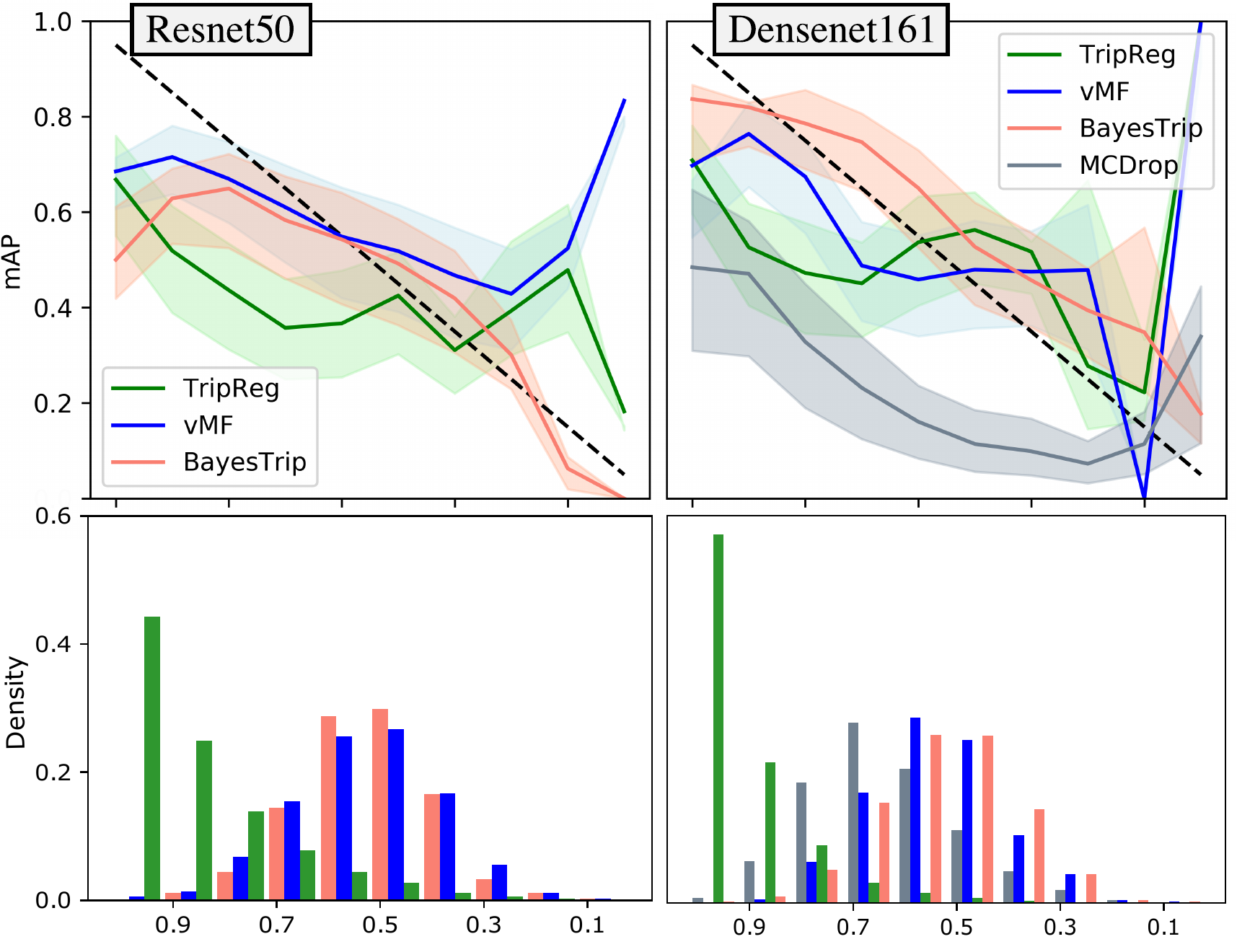}
    \caption{Calibration plots for triplet regression (TripReg), MC dropout (MCDrop), Bayesian triplet loss with Gaussian distribution (BayesTrip) and Bayesian triplet loss with von Mises-Fisher distribution (vMF).
    The solid line indicates mAP@5 and the shaded area covers from mAP@1 to mAP@10. Note how, for both backbones, the Gaussian distributed embeddings are better calibrated, especially for uncertain queries.}
    \vspace{-4mm}
     \label{fig:calibration_cub}
\end{figure}

\subsection{CUB 200-2011} We first evaluate the models' retrieval performance and calibration performance. The CUB 200-2011 dataset~\cite{Wah2011TheCB} consists of $11,788$ images of $200$ bird species. The birds are captured from different perspectives, making this a challenging dataset for image retrieval. We divide the first $100$ classes into the training set and the last $100$ classes into the test set similarly to \citet{musgrave2020metric}. Thus, the trained models have not seen any of the bird species in the test set, and the learned features must generalize well across species.

Table~\ref{tab:retrieval_performance_cub} shows the retrieval performance for the CUB200 dataset for the models with both a Resnet50 and Densenet161 backbone. We notice that the larger backbone improves the retrieval performance across all models. MC dropout performs worse than the other models in terms of predictive performance and uncertainty quantification. 
The triplet loss and the triplet regression have slightly better predictive performance than the proposed Bayesian models, however both of the Bayesian models, especially with the Gaussian embeddings, produce notably better uncertainty estimates. 

To gain a better understanding of which images have high and low variance estimates, Fig.~\ref{fig:most_certain_bird} shows the $12$ queries with highest and lowest variance. The (green/red) border indicates if the image's nearest neighbor is correctly retrieved. The network correctly associates high variance to images where the bird blends in with its surroundings, and low variance to birds that are centered or easily distinguishable by their color or patterns.  

For applications, the user will often be interested in the covariance of the distance $\mathrm{cov}[\Delta]$ between a query and its nearest neighbor. The covariance depends on both the variance of the query and its nearest neighbor. Figure~\ref{fig:bird_confidence} shows six examples of queries and nearest neighbors which exhibit high or low covariances. The model assigns low confidence when multiple birds are present or when a bird blends into its surroundings. High confidence is associated with birds with unique patterns or colors.

Figure~\ref{fig:calibration_cub} shows the Bayesian model with Gaussian embeddings is better calibrated than other methods which produce uncertainty estimates, especially for uncertain queries. The queries are divided into $10$ equi-sized bins. For each bin the mAP@\{1,\,5,\,10\} is calculated, indicated with the top part of the shaded area, the solid line, and lower part of shaded area respectively. The black dotted line shows a perfectly calibrated model.

This experiment shows our proposed Bayesian triplet loss produces retrieval performance comparable to the triplet loss across two strong backbones. Further, the experiment shows that our model produces very well-calibrated uncertainty estimates based on the ECE metric, calibration plots, and qualitative visualizations.

\begin{table*}
\centering
\resizebox{!}{0.105\textwidth}{%
\begin{tabular}{ll|ccccccccc}
                         & & R@1  & R@5  & R@10 & M@1  & M@5  & M@10 &ECE@1			&ECE@5			&ECE@10\\ \hline \hline
\multirow{4}{*}{\rotatebox{90}{ResNet50}} 
&Triplet			& 0.451	&  0.723	&  0.813	&  0.451	& 0.255	&  0.179	&	&	& \\
&TripReg			& 0.447	&0.712	& 0.813	&0.447	&0.254	&0.178	&0.292	&0.479	&0.555 \\
&Bayes vMF		&\bf 0.471&	\bf 0.733&	\bf 0.827&	0\bf.471&	\bf0.270&	\bf 0.191&	0.117&	\bf 0.162&	\bf0.225\\
&Bayes Triplet		&0.431	&0.696	&0.795	&0.431	&0.235	& 0.163	 &\bf 0.094	&  0.271	& 0.343\\ \hline
\multirow{5}{*}{\rotatebox{90}{Densenet161}} 
&Triplet			&\bf 0.495	&\bf 0.751	&\bf 0.837	&\bf 0.495	&\bf 0.293	&\bf 0.212	&	&	& \\
&Triplet (MC=50)			&0.470	&0.717	&0.813	&0.470	&0.269	&0.191	&0.178	&0.367	&0.438 \\
&TripReg	 				&0.481	&0.744	&0.836	&0.481	&0.285	&0.205	&0.263	&0.456	&0.536 \\
&Bayes vMF         &0.478&	0.740&	0.834&	0.474&	0.284&	0.204&	0.160&	0.288&	0.355 \\
&Bayes Triplet		&0.467	&0.724	&0.814	&0.467	&0.269	&0.192	 &\bf0.101	&\bf 0.134	&\bf 0.208\\
\end{tabular}
}
\caption{Recall (R), Mean Average Precision (M) and Expected Calibration Error (ECE) at $1$, $5$ and $10$ on the CAR196 dataset.}
\label{tab:retrieval_performance_car}
\end{table*}

\begin{figure*}
    \centering
    \includegraphics[width=0.9\textwidth]{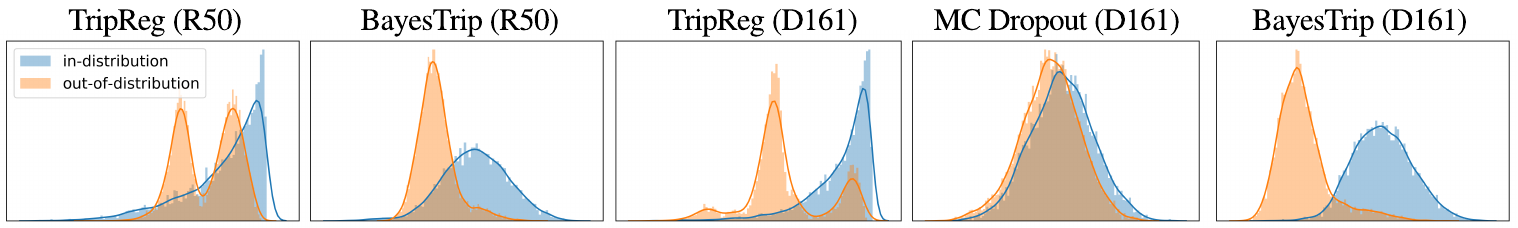}
    \caption{Histogram of the covariance of the distances between queries and their nearest neighbor for triplet regression (TripReg), Bayesian triplet model with Gaussian embeddings (BayesTrip) and MC Dropout for Resnet50 (R50) and Densenet161 (D161). Note that in-distribution (blue) and out-of-distribution (orange) covariances are significantly more separated for the Bayesian model.}
    \vspace{-2mm}
     \label{fig:OODvsID}
\end{figure*}

\begin{figure}
     \centering
     \includegraphics[width=\hsize]{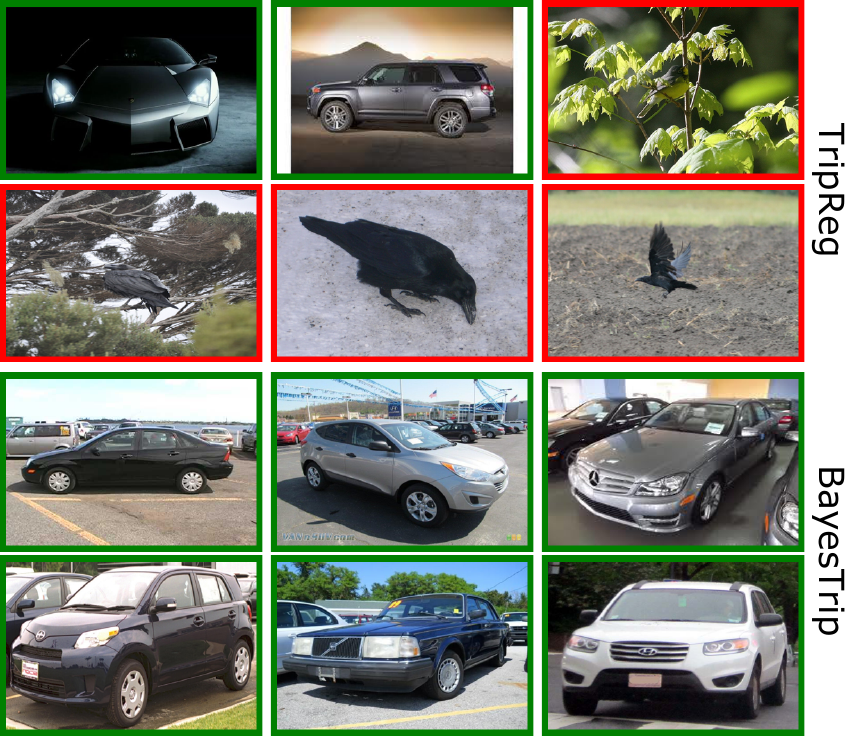}
     \caption{Least probable among in and out of distribution for triplet regression (TripReg) and Bayesian triplet Loss with Gaussian embeddings (BayesTrip). Cars are out of distribution queries and birds are in distribution queries.}
     \vspace{-4mm}
     \label{fig:least_certain_tr}
\end{figure}

\subsection{Stanford Car-196} Next, we investigate how well the models perform on out-of-distribution (OOD) examples. This is an important capability for image retrieval systems since it is infeasible to retrain the models continuously, and in many practical applications unseen categories are likely to be added to the database or used as queries over time. To test the OOD capabilities of the models, we use the Stanford Car-196~\cite{6755945} dataset and the models trained on the CUB200 dataset. The Car-196 dataset is composed of $16,185$ images of $196$ classes of cars. It is traditionally a classification dataset, but can be cast as a retrieval dataset by using the first $98$ categories as a training set and the last $98$ ones as a test set~\cite{musgrave2020metric}.

First, we evaluate how well the models generalize to the Car-196 test set. Table~\ref{tab:retrieval_performance_car} shows that both Bayesian models match (and for vMF embeddings with the ResNet50 backbone surpass) the retrieval performance of the triplet loss. The ECE metric shows that the Bayesian models are significantly better calibrated. Among the Bayesian models, the Gaussian embeddings give better uncertainty estimates, whereas the vMF embeddings give slightly better predictive performance.


\begin{figure*}
     \centering
    \includegraphics[width=0.9\hsize]{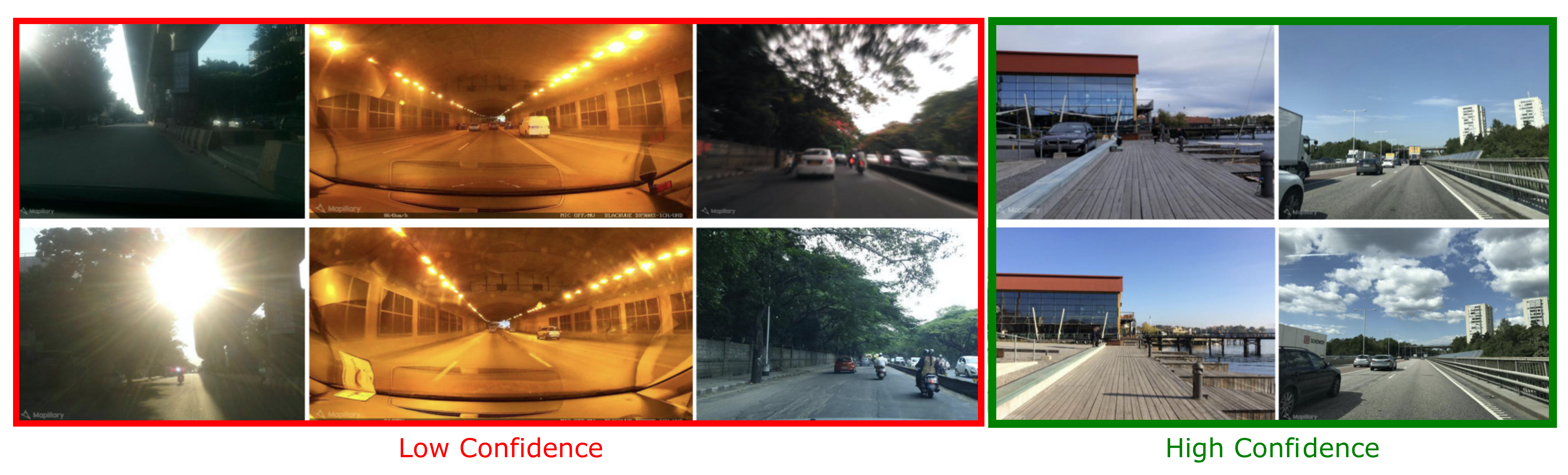}
    \vspace{-0.5cm}
     \caption{Low-confidence (columns 1--3) and high-confidence (columns 4 and 5) retrievals from the MSLS dataset. The top row shows queries and the bottom row shows their NNs. Our model gives low confidence to images with harsh sunlight, blur and ambiguous tunnels and vegetation. In contrast, high confidence is given for landmark buildings.}
     \label{fig:msls_confidence}
\end{figure*}

\begin{table*}
\centering
\resizebox{!}{0.105\textwidth}{%
\begin{tabular}{ll|ccccccccc}
                         & & R@1  & R@5  & R@10 & M@1  & M@5  & M@10 &ECE@1			&ECE@5			&ECE@10\\ \hline \hline
\multirow{4}{*}{\rotatebox{90}{ResNet50}} 
&Triplet			&   0.350&	    0.495&	    0.554&      0.350&	    0.235&	    0.210&	&	&	\\
&TripReg			&   0.349& \bf	0.499&	    0.551&	    0.349&	    0.238&	    0.215&	 0.482&	0.571&	0.593\\
&Bayes vMF		&   0.349& 	    0.494&	\bf 0.559&	    0.349&	\bf 0.249&	\bf 0.226&	 0.482&	0.571&	0.593\\
&Bayes Triplet		&\bf0.354&      0.489&      0.549&\bf   0.354&      0.241&      0.218&\bf0.208&\bf0.319&\bf0.341\\ \hline 
\multirow{5}{*}{\rotatebox{90}{Densenet161}} 
&Triplet			& \bf 0.386&	\bf 0.531&	 0.583&	\bf 0.386&	\bf 0.270&	\bf 0.246&	&	&		\\
&Triplet (MC=50)			&0.282&	0.412&	0.458&	0.282&	0.188&	0.163&	0.540&	0.625&	0.648	\\
&TripReg	 				& \bf 0.386&	0.529&	\bf 0.596&	\bf 0.386&	0.268&	0.245&	0.424&	0.543&	0.566\\
&Bayes vMF & 0.383&	0.526&	0.588&	0.383&	0.268&	0.245&	0.227&	0.327&	0.350 \\
&Bayes Triplet		&0.364&	0.506&	0.571&	0.364&	0.252&	0.228&	\bf 0.196&	\bf 0.264&	\bf 0.283\\
\end{tabular}
}
\caption{Recall (R), Mean Average Precision (M) and Expected Calibration Error (ECE) at $1$, $5$ and $10$ on the MSLS dataset.}
\vspace{-0.4cm}
\label{tab:retrieval_performance_MSLS}
\end{table*}

Second, we show that the Bayesian models can detect OOD queries. To do so, we construct a database consisting of bird images. We retrieve the nearest neighbor with bird queries -- in distribution (ID) -- and car queries -- out-of-distribution (OOD). We expect the distance to the nearest neighbor of the OOD queries to have high uncertainty compared to ID queries. Figure~\ref{fig:OODvsID} shows that the Bayesian model with Gaussian embeddings is significantly better at differentiating between ID and OOD queries. 

The six least confident Gaussian query embeddings from our Bayesian model (Fig.~\ref{fig:least_certain_tr}) reveal OOD queries (cars), whereas for the triplet regression, several ID queries (birds) are seen among the least certain. Even though these queries are challenging birds, we would expect the model to be less confident about images from a completely different domain.

This experiment shows that the confidences generated by the Bayesian models generalize well to out-of-distribution examples and can be used to discriminate between queries from in- and out-of-distribution. Again, the Bayesian models match the predictive performance to the other methods, and achieve state-of-the-art uncertainty estimates.

\subsection{Mapillary Street Level Sequences (MSLS)}
Finally, we show that the Bayesian models also have competitive retrieval performance and state-of-the-art uncertainty estimates for large datasets. MSLS~\cite{Warburg_CVPR_2020} is the largest and most diverse place recognition dataset currently available, and comprises $1.6M$ images from $30$ cities spanning six continents. The goal in place recognition is to retrieve images from the same place as a query image (where the same place is typically defined within a radius of $25$ m). This is challenging due to the large number of unique places and large spectrum of appearance changes for each place, such as weather, dynamic, structural, view-point, seasonal and day/night changes. We use the train/test split recommended in \cite{Warburg_CVPR_2020}, training on $24$ cities and testing on six other different cities.

The Bayesian models achieve a performance comparable to the traditional triplet loss, even outperforming it when using the von Mises-Fisher distribution (Table~\ref{tab:retrieval_performance_MSLS}). Furthermore, the Bayesian model with Gaussian embeddings provides state-of-the-art uncertainty estimates for both backbones.

Figure~\ref{fig:msls_confidence} illustrates how the Bayesian model with Gaussian embeddings associates low confidence to images that are difficult to retrieve due to harsh sunlight, blur or the ambiguous, repetitive patterns in tunnels. Furthermore, the model is able to assign high confidence to images that have unique structural appearance, as seen in the last two columns. This experiment shows that the Bayesian models have a competitive performance in very large and challenging datasets, matching the predictive performance of the standard triplet loss, and producing state-of-the-art uncertainty estimates.

\subsection{Trade-off between predictive performance and uncertainty quantification}

In many applications, reliable uncertainties are a requirement. This is common for method that needs to provide guarantees of certifiability. In safety-critical applications uncertainty is important as it can ensure timely interventions from the user e.g. when image retrieval is used in loop closure for robot localization.

Contemporary probabilistic methods often exhibit a slight decrease in performance over non-probabilistic methods, but the trade-off is worth making in the above-mentioned examples. We indeed observe a similar trend, where the Bayesian Triplet Loss is either on par or slightly below state-of-the-art in predictive performance, but is clear state-of-the-art in terms of uncertainty quantification. One reason for the decrease in predictive performance is that we have less free parameters for the mean prediction than other methods: To ensure a fair comparison, we constraint the different models to have the same number of parameters, which imply that we use some of our capacity on predicting $\sigma$.  This lessens our capacity for mean predictions.

\section{Conclusion}

We have proposed to model image embeddings as stochastic features rather than point estimates. We derive a new likelihood that follows the intuition of the triplet loss, but works for stochastic features. We introduce a prior over the feature space that together with our likelihood enables us to learn either Gaussian distributed or von Mises-Fisher distributed stochastic features. The proposed method, the Bayesian triplet loss, produces state-of-the-art uncertainty estimates, without sacrificing predictive performance compared to the triplet loss.
Quantification of uncertainty in image retrieval is vital for safety-critical applications, while reliable uncertainty estimates also open many other doors, for example related to interpretability or user-friendly retrieval interfaces. We speculate that reliable uncertainty estimates can also be used for hard negative mining and avoidance of outliers in query expansion~\cite{queryexpansion}. 

\vspace{2mm}
\begin{spacing}{0.25}
\begin{footnotesize}
\textbf{Acknowledgments.}
This work was supported in part by a research grant (15334) from VILLUM FONDEN. This project has received funding from the European Research Council (ERC) under the European Union's Horizon 2020 research and innovation programme (grant agreement n\textsuperscript{o} 757360), from the Spanish Government
(PGC2018-096367-B-I00) and the Aragon Government (DGA T45 17R/FSE). MJ is supported by a research grant from the Carlsberg foundation (CF20-0370).
\end{footnotesize}
\end{spacing}

{\small
\bibliographystyle{plainnat}
\bibliography{ms}
}

\end{document}




\maketitle

\section{Introduction}

In this supplementary material we give additional details into the derivations of the mean and the variance for the Gaussian approximation of the proposed likelihood. We show through simulations that the Gaussian approximation of the likelihood is very accurate, even for a small number of dimensions. We provide the full derivation of the Expected Lower Bound (ELBO). Finally, we provide sample code for the proposed Bayesian Triplet loss, which highlights that the proposed method is simple to implement. We will make the entire code available upon acceptance.

\section{Gaussian Likelihood}

First, we provide the details into the derivations of the first two moments of the proposed Gaussian approximation of the likelihood.

\subsection{The Mean}

The expression
\begin{align}
  \mathbb{E}[\tau]
    &= \mathbb{E}[p(p - 2a)] - \mathbb{E}[n(n - 2a)]
\end{align}
consists of two means of the same form. We first evaluate
\begin{align}
  \mathbb{E}[p(p - 2a)]
    &= \mathbb{E}[p^2] - 2\mathbb{E}[ap]
     = \mathbb{E}[p^2] - 2\mathbb{E}[a]\mathbb{E}[p],
\end{align}
where we have used that the expectation of the product of two independent variables
is the product of the expectations. Using identical arguments for the second term of $\mathbb{E}[\tau]$ we get
\begin{align}
  \mathbb{E}[\tau]
    &= \mathbb{E}[p^2] - 2\mathbb{E}[a]\mathbb{E}[p]
     - \mathbb{E}[n^2] + 2\mathbb{E}[a]\mathbb{E}[n] \\
    &= \mathbb{E}[p^2] - \mathbb{E}[n^2]
     - 2\mathbb{E}[a](\mathbb{E}[p] - \mathbb{E}[n]) \\
    &= \mu_p^2 + \sigma_p^2 - \mu_n^2 - \sigma_n^2
     - 2\mu_a(\mu_p - \mu_n).
    \end{align}

\subsection{The variance}
The expression 
\begin{align}
  \mathrm{var}[\tau]
    &= \mathrm{var}[p(p - 2a) - n(n - 2a)] \\
    &= \mathrm{var}[p(p - 2a)] + \mathrm{var}[n(n - 2a)] - 2  \mathrm{cov}[p(p - 2a), n(n - 2a)]
\end{align}
consists of three terms. We treat the terms one by one
\begin{align}
  \mathrm{var}[p(p - 2a)]
    &= \mathrm{var}[p^2 - 2ap] \\
    &= \mathrm{var}[p^2] + 4\mathrm{var}[ap] - 4\mathrm{cov}[p^2, ap] \\
    \begin{split}
    &= 2\sigma_p^4 + 4\mu_p^2 \sigma_p^2
     + 4\left( \mathrm{var}[a] + \mathbb{E}[a]^2 \right)
        \left( \mathrm{var}[p] + \mathbb{E}[p]^2 \right) \\
    &- 4\mathbb{E}[a]^2  \mathbb{E}[p]^2
     - 4\mathrm{cov}[p^2, ap]
    \end{split} \\
    \begin{split}
    &= 2\sigma_p^4 + 4\mu_p^2 \sigma_p^2
     + 4\left( \sigma_a^2 + \mu_a^2 \right) \left( \sigma_p^2 + \mu_p^2 \right)
     - 4\mu_a^2 \mu_p^2 \\
    &- 4\mathrm{cov}[p^2, ap]
    \end{split} 
\end{align}

The last covariance term is

\begin{align}
  \mathrm{cov}[p^2, ap]
    &= \mathbb{E}[ap^3] - \mathbb{E}[ap]E[p^2] \\
    &= \mathbb{E}[a]\mathbb{E}[p^3] - \mathbb{E}[a]\mathbb{E}[p]\mathbb{E}[p^2] \\
    &= \mu_a(\mu_p^3 + 3\mu_p\sigma_p^2) - \mu_a \mu_p(\mu_p^2 + \sigma_p^2) \\
    &= \mu_a \left[\mu_p^3 + 3\mu_p\sigma_p^2 - \mu_p^3 - \mu_p \sigma_p^2) \right] \\
    &= 2 \mu_a \mu_p\sigma_p^2,
\end{align}
such that
\begin{align}
  \mathrm{var}[p(p - 2a)]
    &= 2\Big[ \sigma_p^4 + 2\mu_p^2 \sigma_p^2
     + 2\left( \sigma_a^2 + \mu_a^2 \right) \left( \sigma_p^2 + \mu_p^2 \right)
     - 2\mu_a^2 \mu_p^2
     - 4 \mu_a \mu_p\sigma_p^2 \Big].
\end{align}

The second term of the variance follows the same structure,
\begin{align}
  \begin{split}
  \mathrm{var}[n(n - 2a)]
    &= 2\Big[ \sigma_n^4 + 2\mu_n^2 \sigma_n^2
     + 2\left( \sigma_a^2 + \mu_a^2 \right) \left( \sigma_n^2 + \mu_n^2 \right)
     - 2\mu_a^2 \mu_n^2
     - 4 \mu_a \mu_n\sigma_n^2 \Big].
  \end{split}
\end{align}

The last term of the variance is
\begin{align}
  \mathrm{cov}[p(p - 2a), n(n - 2a)]
    &= \mathrm{cov}[p^2 - 2ap, n^2 - 2an] \\
    &= \mathrm{cov}[p^2, n^2] 
     -2\mathrm{cov}[p^2, an]
     -2\mathrm{cov}[ap, n^2] 
     +4\mathrm{cov}[ap, an] \\
    &= 4\mathrm{cov}[ap, an] \\
    &= 4\mathbb{E}[a^2pn] - 4\mathbb{E}[ap]\mathbb{E}[an] \\
    &= 4\mathbb{E}[a^2]\mathbb{E}[p]\mathbb{E}[n] - 4\mathbb{E}[a]^2\mathbb{E}[p]\mathbb{E}[n] \\
    &= 4\mathbb{E}[p]\mathbb{E}[n]\left( \mathbb{E}[a^2] - \mathbb{E}[a]^2 \right) \\
    &= 4\mathbb{E}[p]\mathbb{E}[n]\mathrm{var}[a] \\
    &= 4\mu_p\mu_n\sigma_a^2.
\end{align}

Thus, given a mean and variance estimate for each embedding, we can analytically find the mean and variance of the Gaussian that approximate our proposed likelihood. In the next section, we show experimentally that this is good approximation. 
    
\subsection{Simulation Approximation of Gaussian Likelihood}

In this section, we show that the Gaussian approximation of our proposed likelihood is accurate even for low dimensions. We sample $\mu_a$, $\mu_p$ and $\mu_n$ from a $D$-dimensional standard Gaussian, and $\sigma^2_a$, $\sigma^2_p$ and $\sigma^2_n$ from a $1$-dimensional standard Gaussian, where we ensure that $\sigma^2_a$, $\sigma^2_p$ and $\sigma^2_n$ are positive by taking the absolute value of the sample. We then generate $N$ samples from $a \sim N(\mu_a, \sigma_a^2)$, $p \sim N(\mu_p, \sigma_p^2)$ and $n \sim N(\mu_n, \sigma_n^2)$, and calculate $\tau = \|a - p \|^2 - \|a - n\|^2$. We plot a histogram over the samples of $\tau$ as a Monte Carlo approximation of the true CDF (histograms in the figures below). We use the means and variances of $a$, $p$ and $n$ to analytically calculate the CDF of our Gaussian approximation (black line in the figures below). The figures show that the approximation is very accurate even for a very low number of dimensions (the deviation between the analytical CDF and the Monte Carlo simulation is only significant for $D < 4$). This means that the approximation is accurate for any practical retrieval system.

\begin{figure}[htp!]
    \centering
     \begin{subfigure}[b]{0.19\linewidth}
         \centering
         \includegraphics[width=\linewidth]{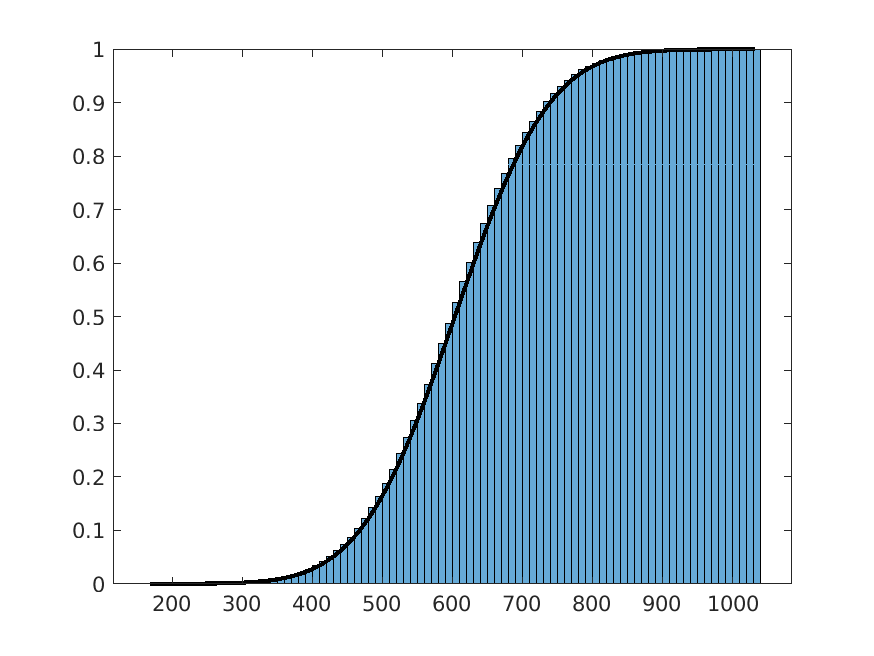} 
     \end{subfigure}
     \begin{subfigure}[b]{0.19\linewidth}
         \centering
         \includegraphics[width=\linewidth]{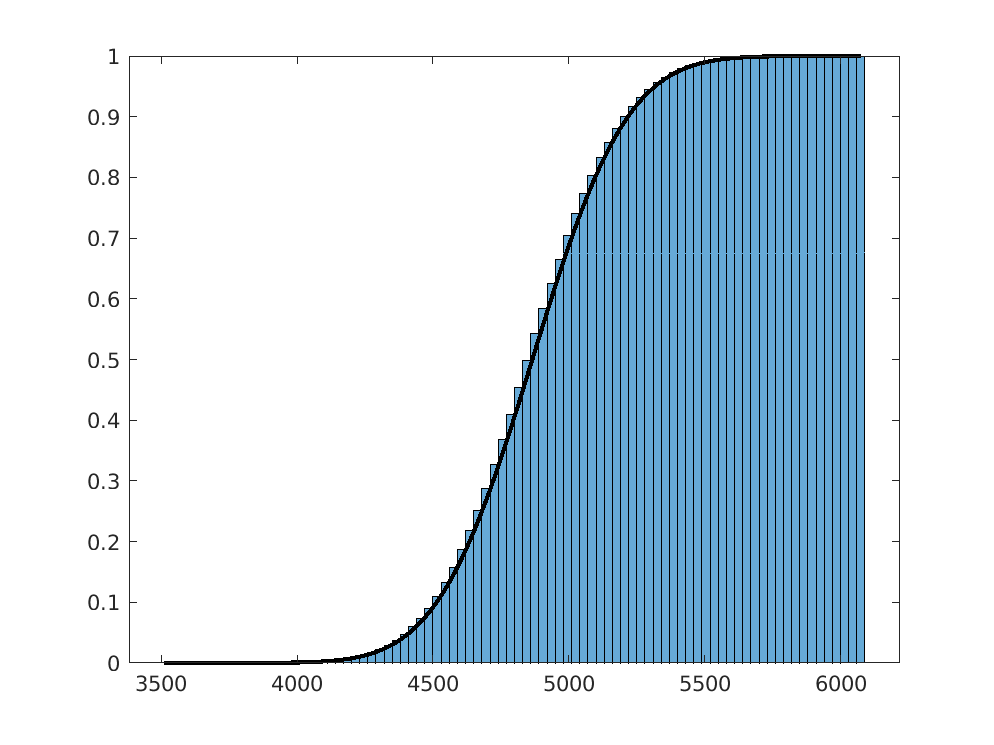}
     \end{subfigure}
     \begin{subfigure}[b]{0.19\linewidth}
         \centering
         \includegraphics[width=\linewidth]{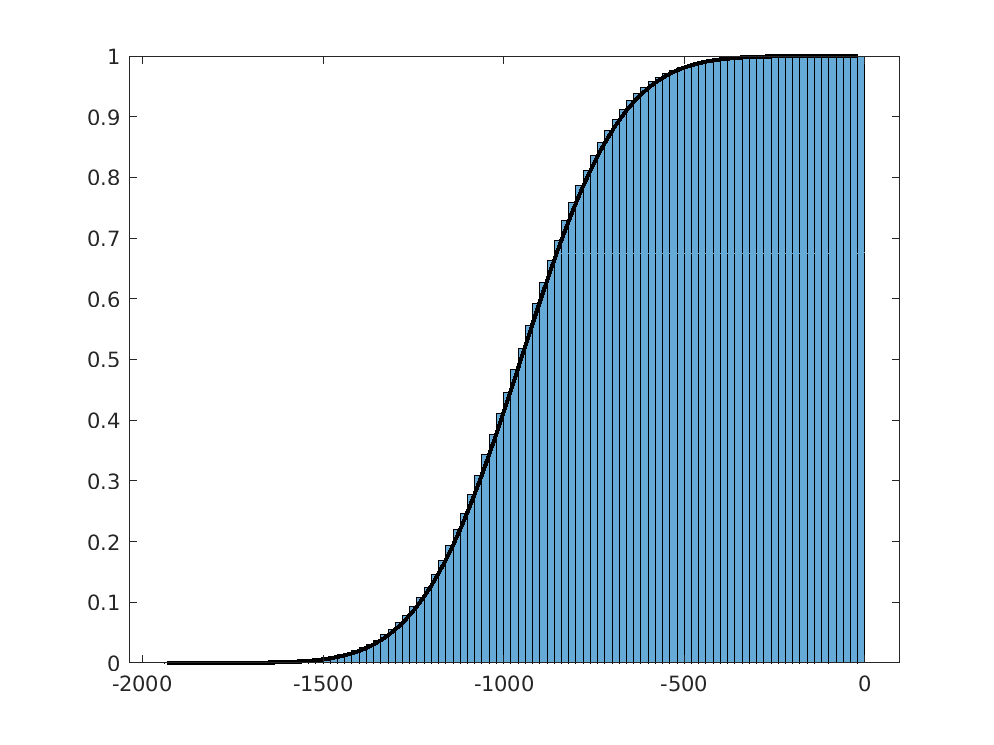}
     \end{subfigure}
     \begin{subfigure}[b]{0.19\linewidth}
         \centering
         \includegraphics[width=\linewidth]{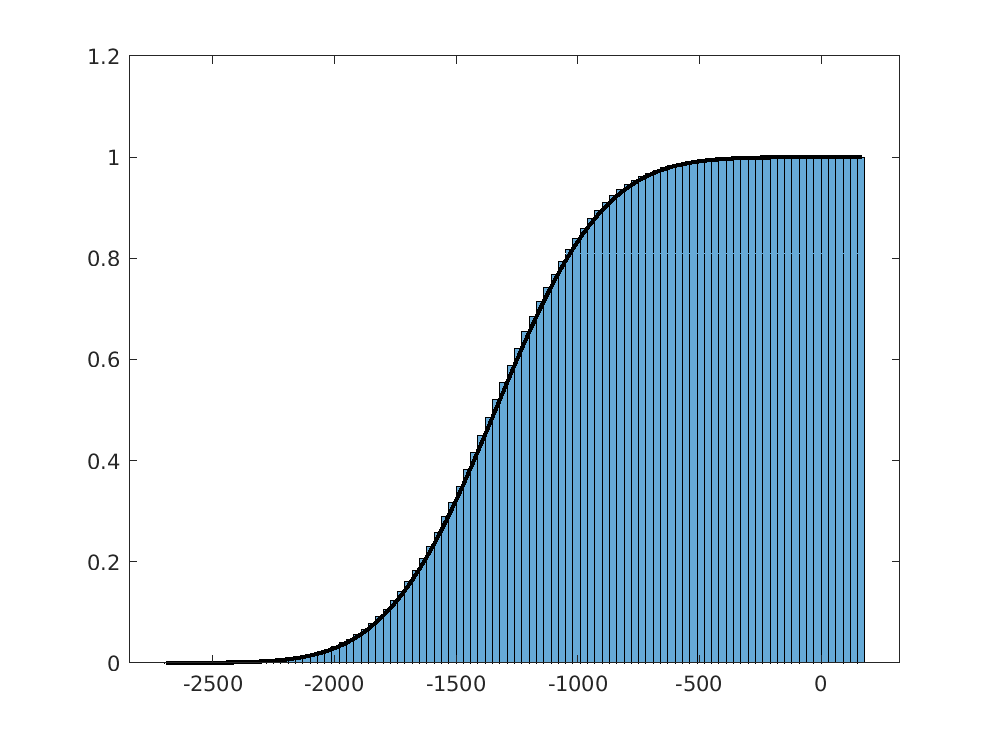}
     \end{subfigure}
     \begin{subfigure}[b]{0.19\linewidth}
         \centering
         \includegraphics[width=\linewidth]{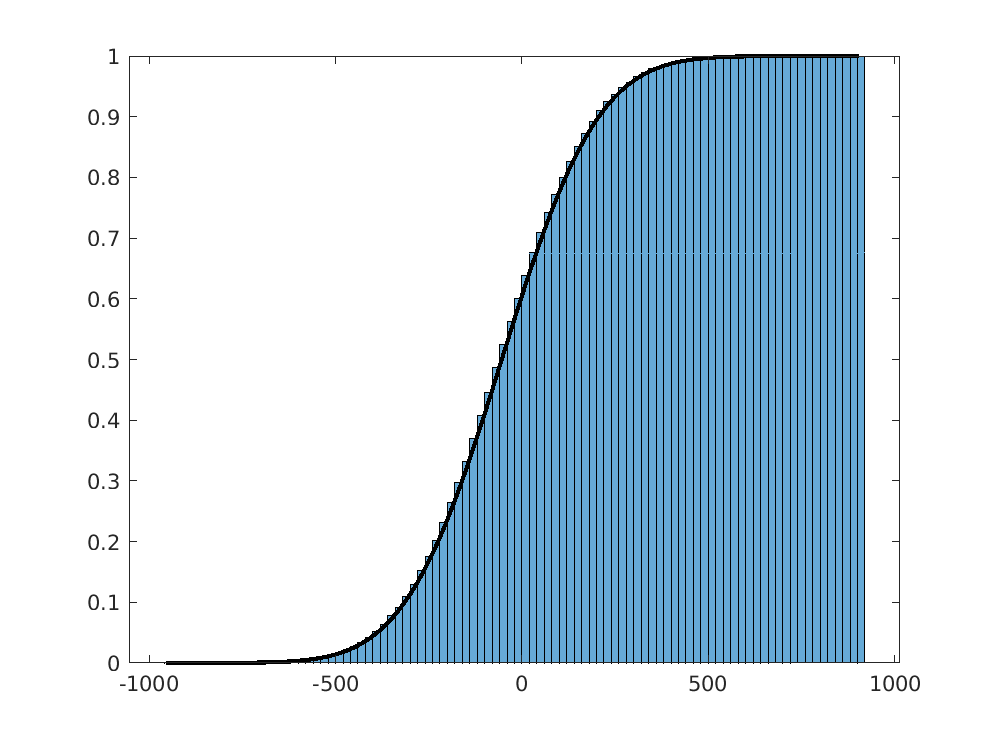}
     \end{subfigure}
     \caption{$D = 2048$}
\end{figure}
\begin{figure}[htp!]
    \centering
     \begin{subfigure}[b]{0.19\linewidth}
         \centering
         \includegraphics[width=\linewidth]{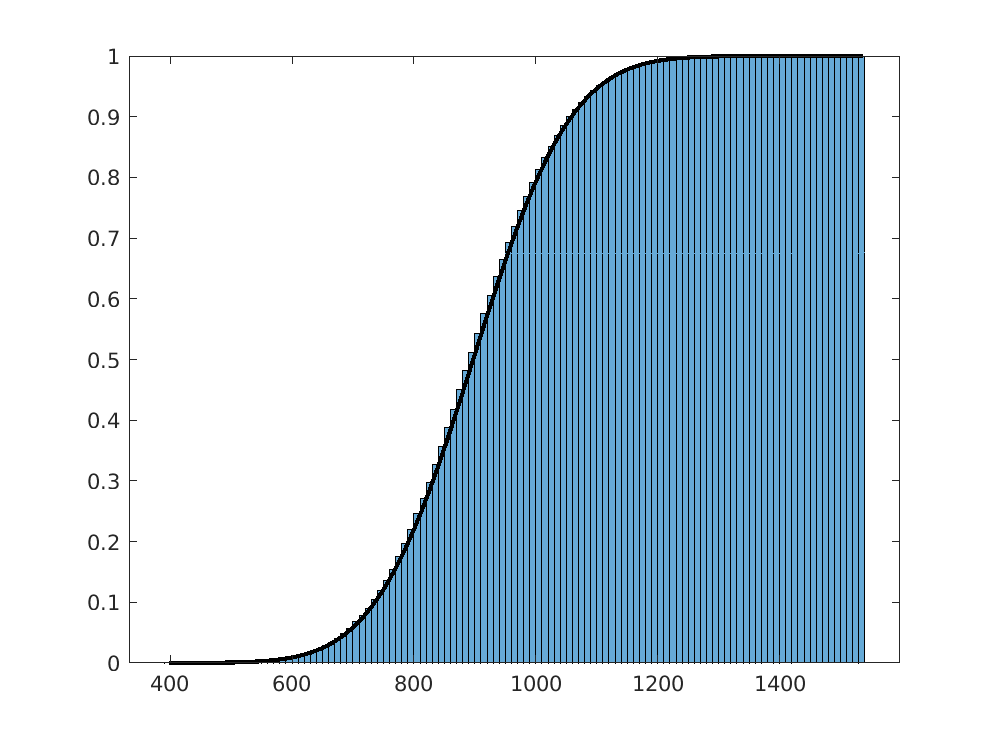} 
     \end{subfigure}
     \begin{subfigure}[b]{0.19\linewidth}
         \centering
         \includegraphics[width=\linewidth]{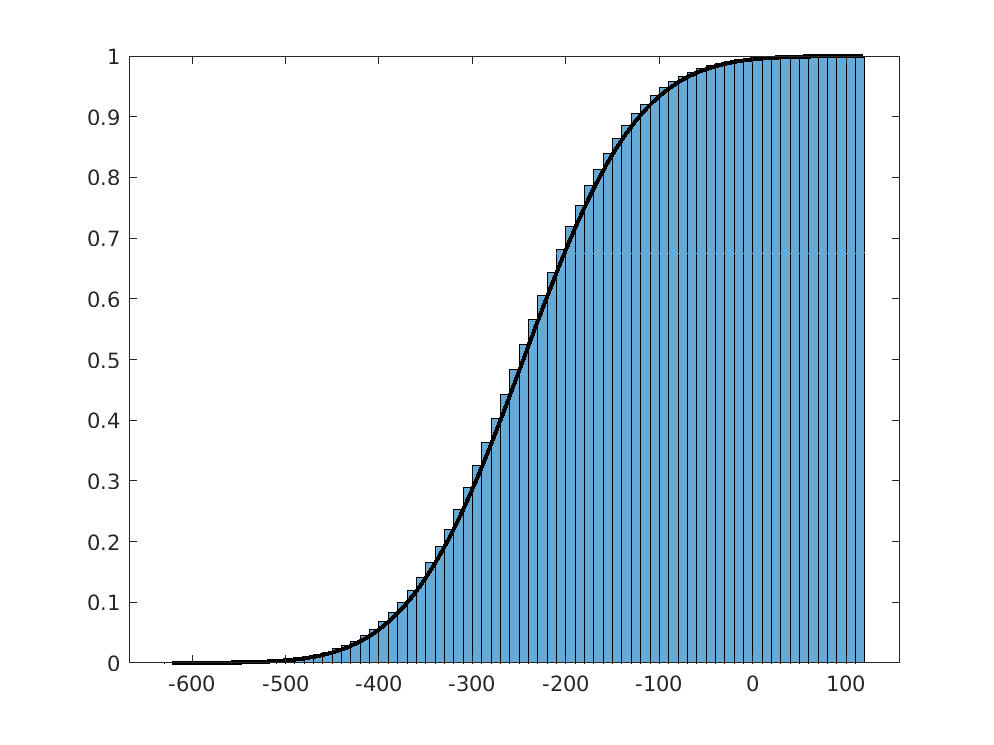}
     \end{subfigure}
     \begin{subfigure}[b]{0.19\linewidth}
         \centering
         \includegraphics[width=\linewidth]{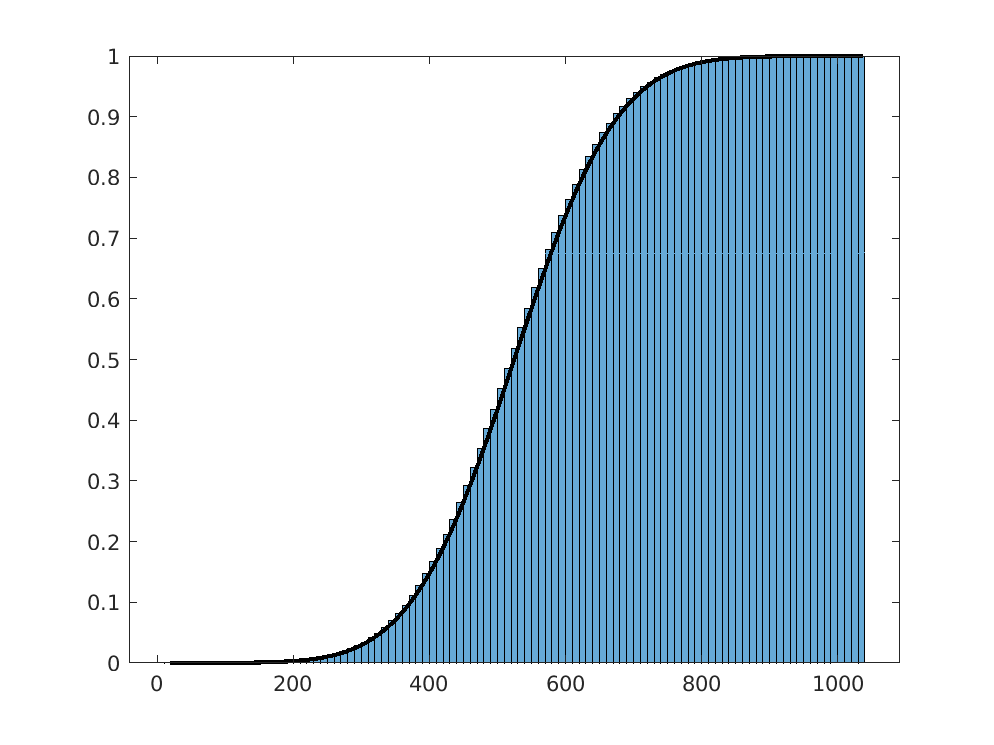}
     \end{subfigure}
     \begin{subfigure}[b]{0.19\linewidth}
         \centering
         \includegraphics[width=\linewidth]{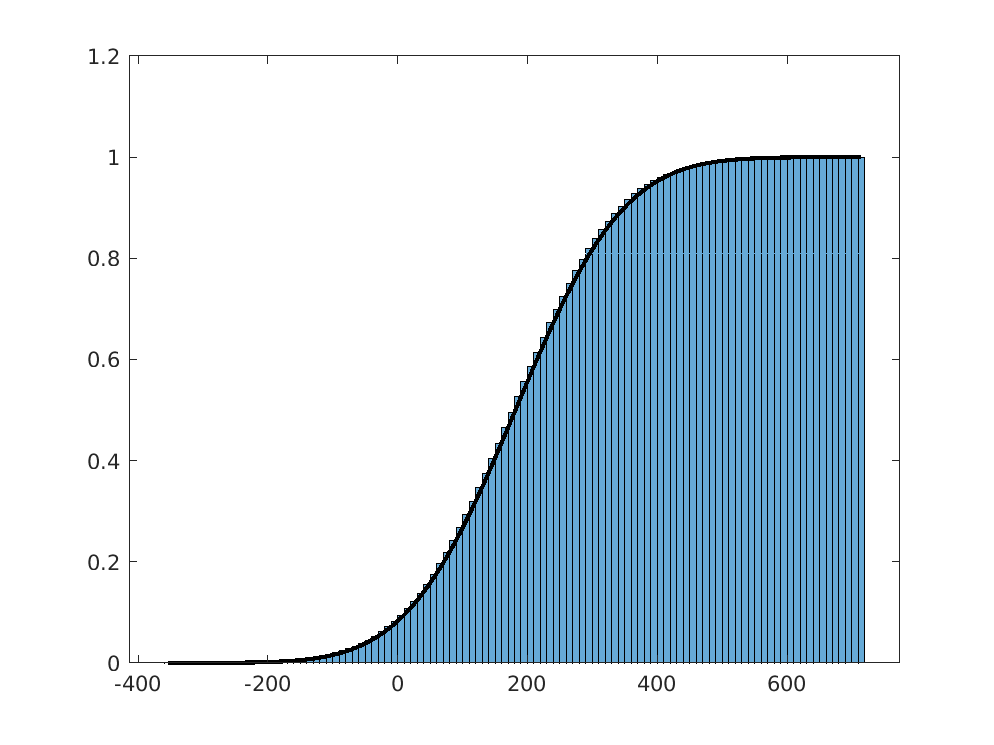}
     \end{subfigure}
     \begin{subfigure}[b]{0.19\linewidth}
         \centering
         \includegraphics[width=\linewidth]{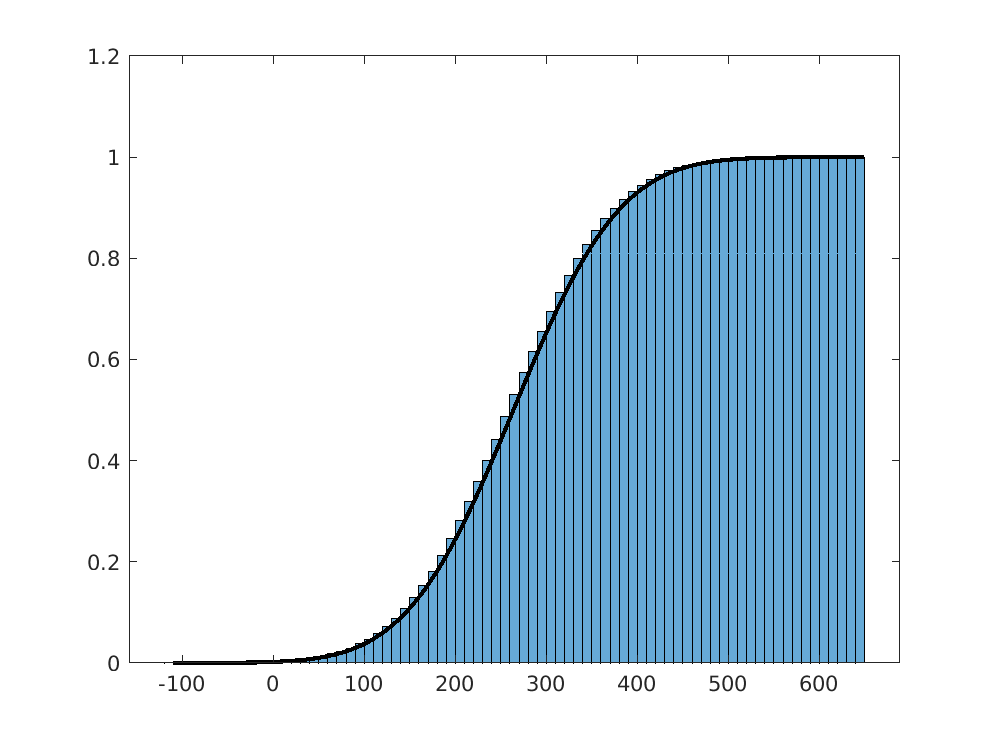}
     \end{subfigure}
     \caption{$D = 512$}
\end{figure}
\begin{figure}[htp!]
    \centering
     \begin{subfigure}[b]{0.19\linewidth}
         \centering
         \includegraphics[width=\linewidth]{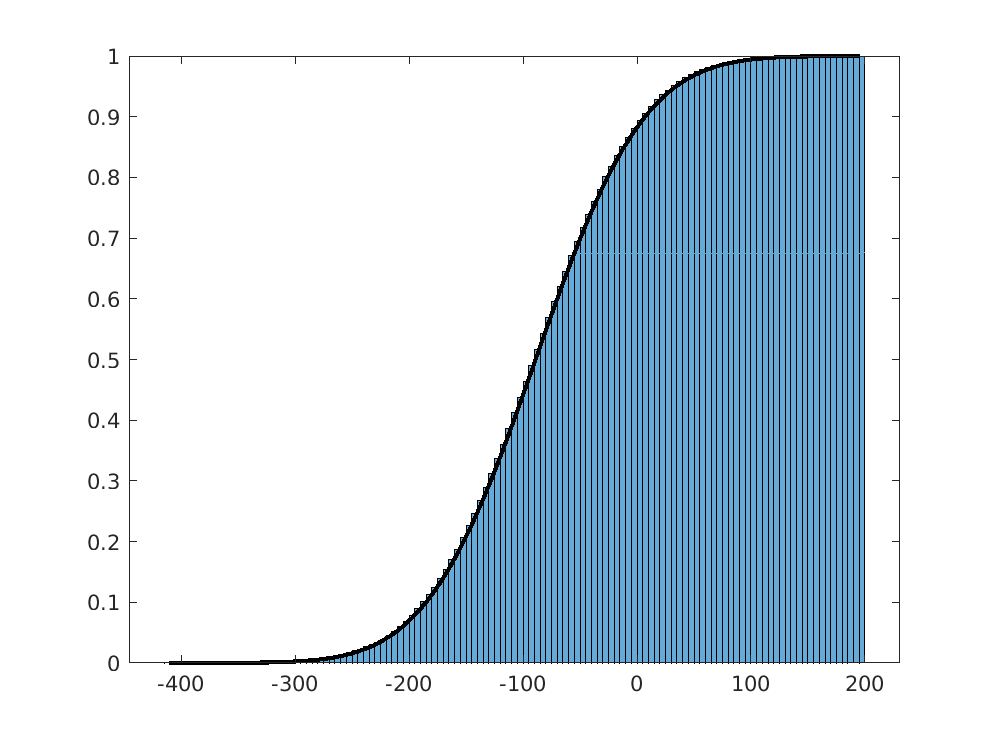} 
     \end{subfigure}
     \begin{subfigure}[b]{0.19\linewidth}
         \centering
         \includegraphics[width=\linewidth]{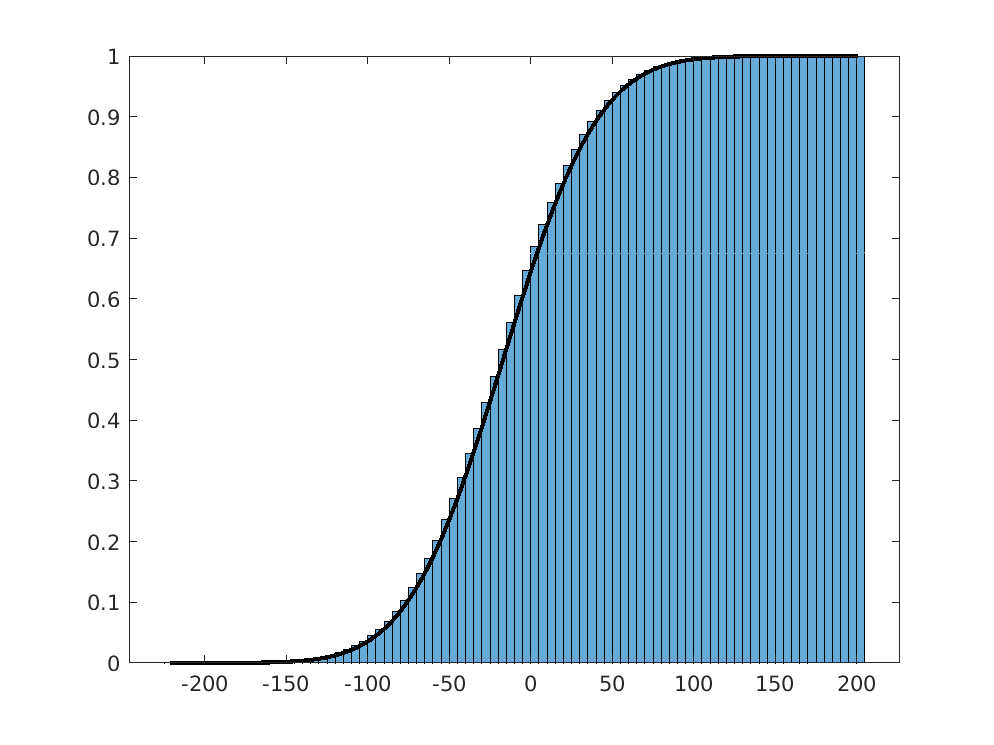}
     \end{subfigure}
     \begin{subfigure}[b]{0.19\linewidth}
         \centering
         \includegraphics[width=\linewidth]{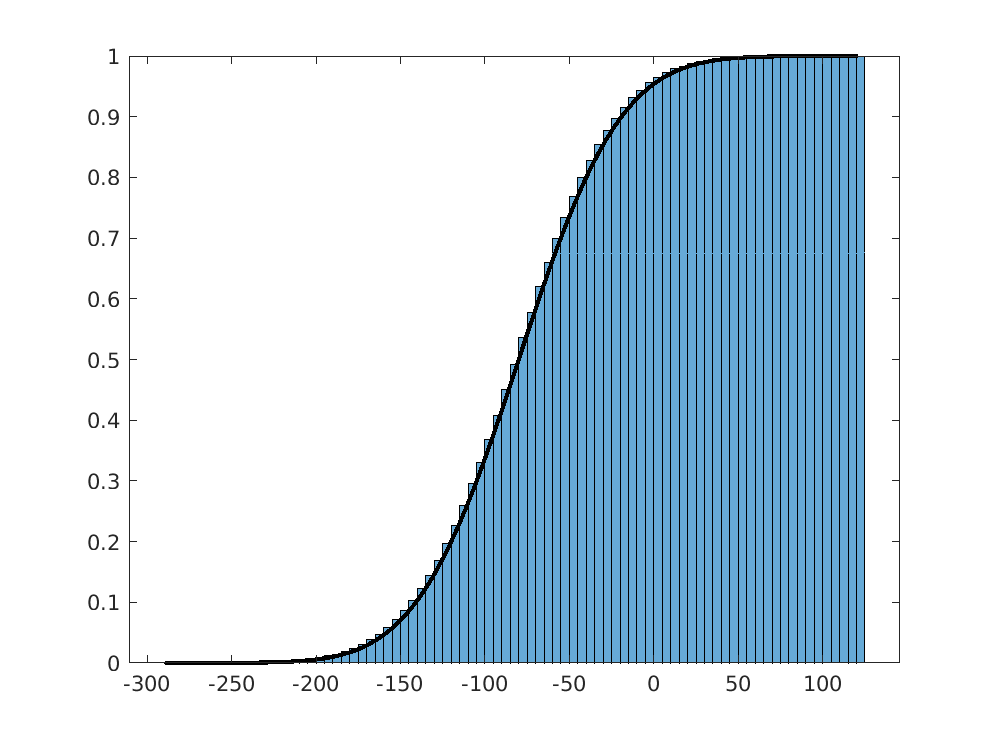}
     \end{subfigure}
     \begin{subfigure}[b]{0.19\linewidth}
         \centering
         \includegraphics[width=\linewidth]{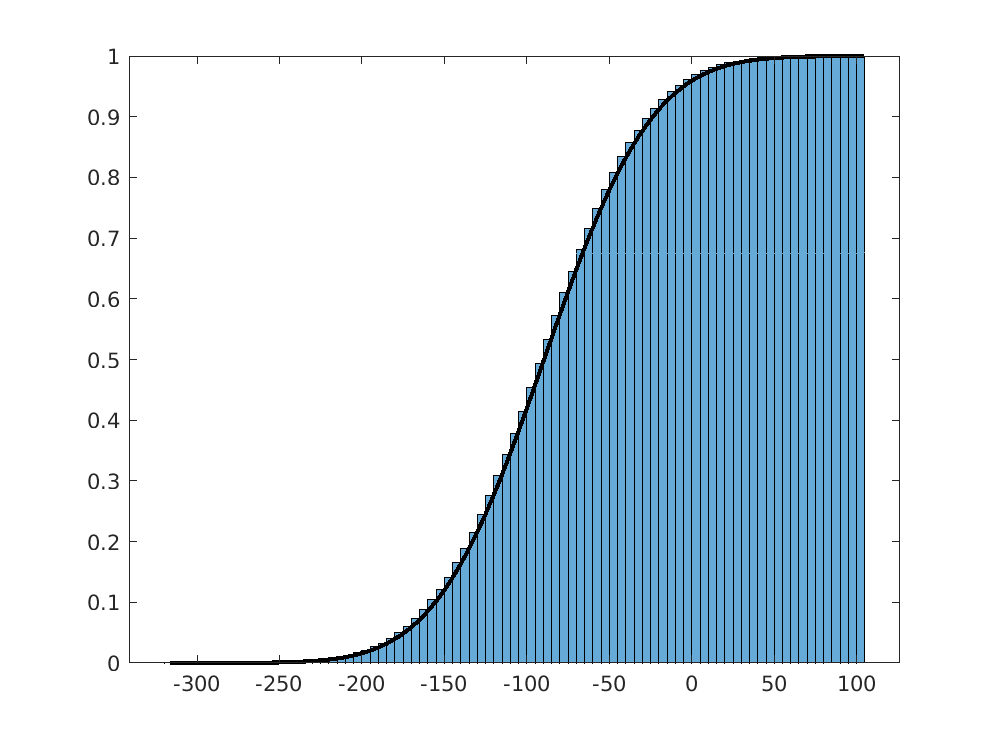}
     \end{subfigure}
     \begin{subfigure}[b]{0.19\linewidth}
         \centering
         \includegraphics[width=\linewidth]{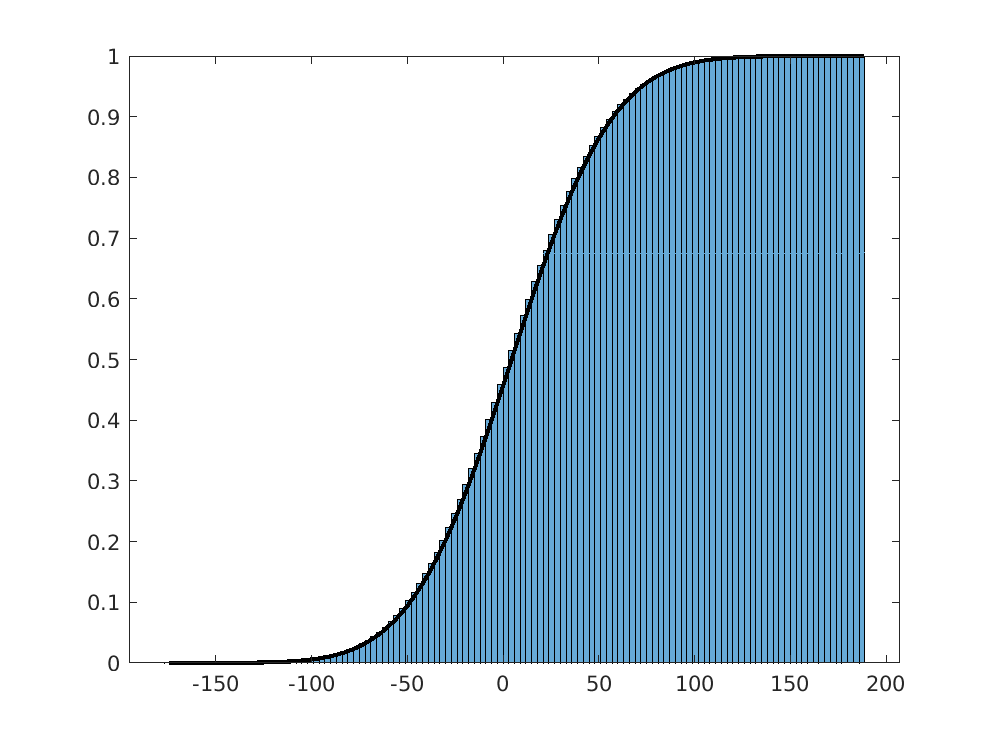}
     \end{subfigure}
     \caption{$D = 128$}
\end{figure}
\begin{figure}[htp!]
    \centering
     \begin{subfigure}[b]{0.19\linewidth}
         \centering
         \includegraphics[width=\linewidth]{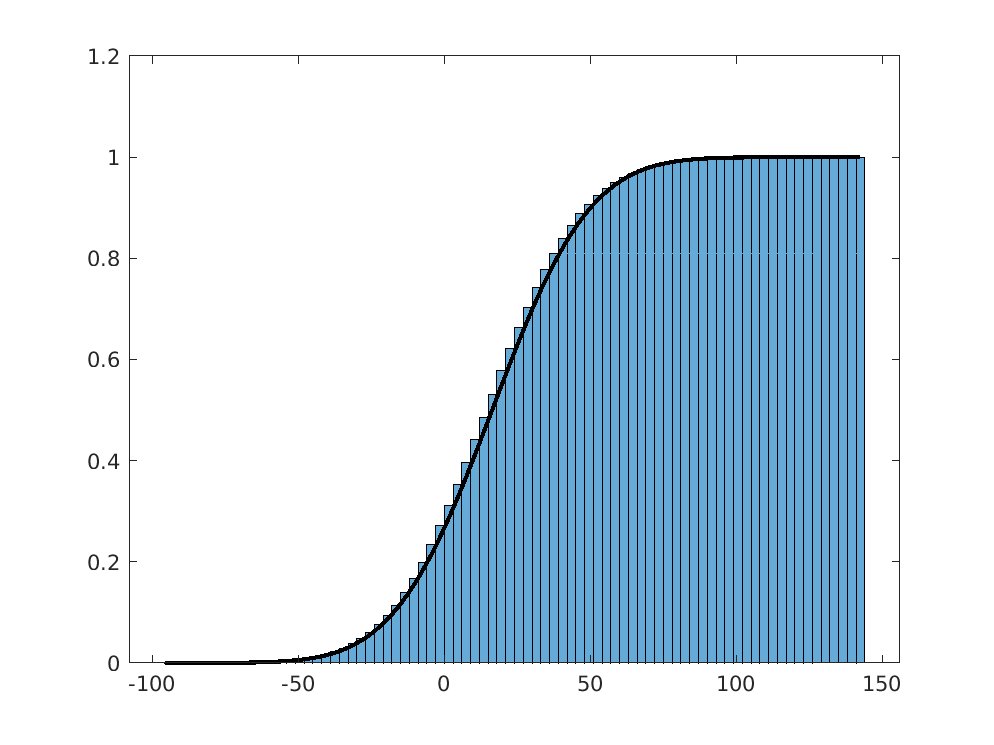} 
     \end{subfigure}
     \begin{subfigure}[b]{0.19\linewidth}
         \centering
         \includegraphics[width=\linewidth]{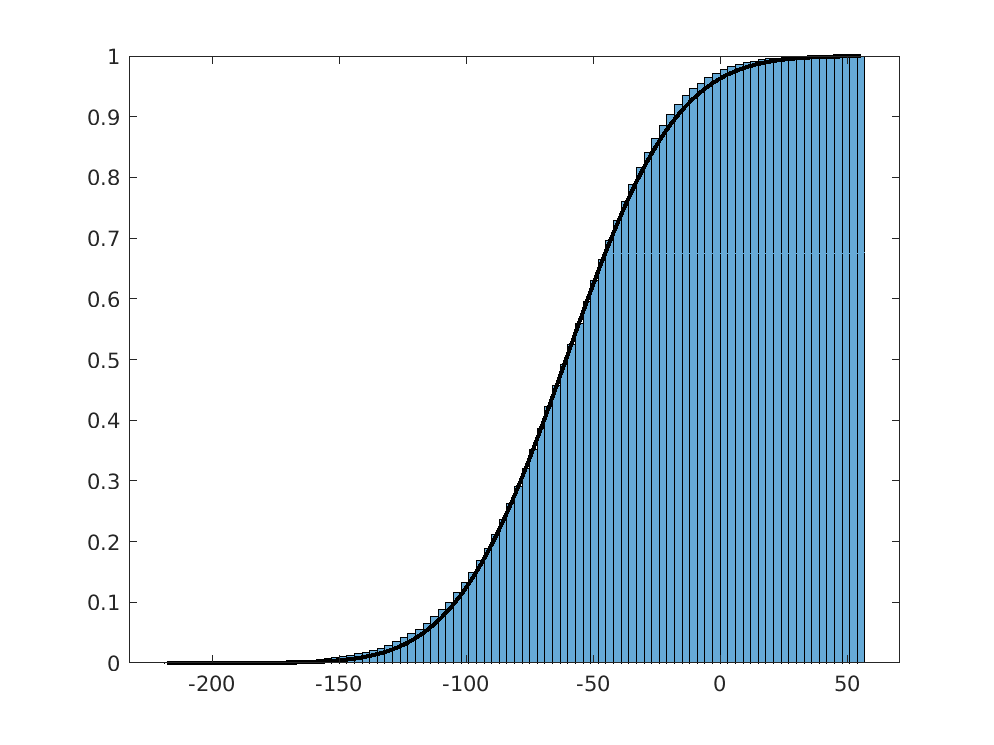}
     \end{subfigure}
     \begin{subfigure}[b]{0.19\linewidth}
         \centering
         \includegraphics[width=\linewidth]{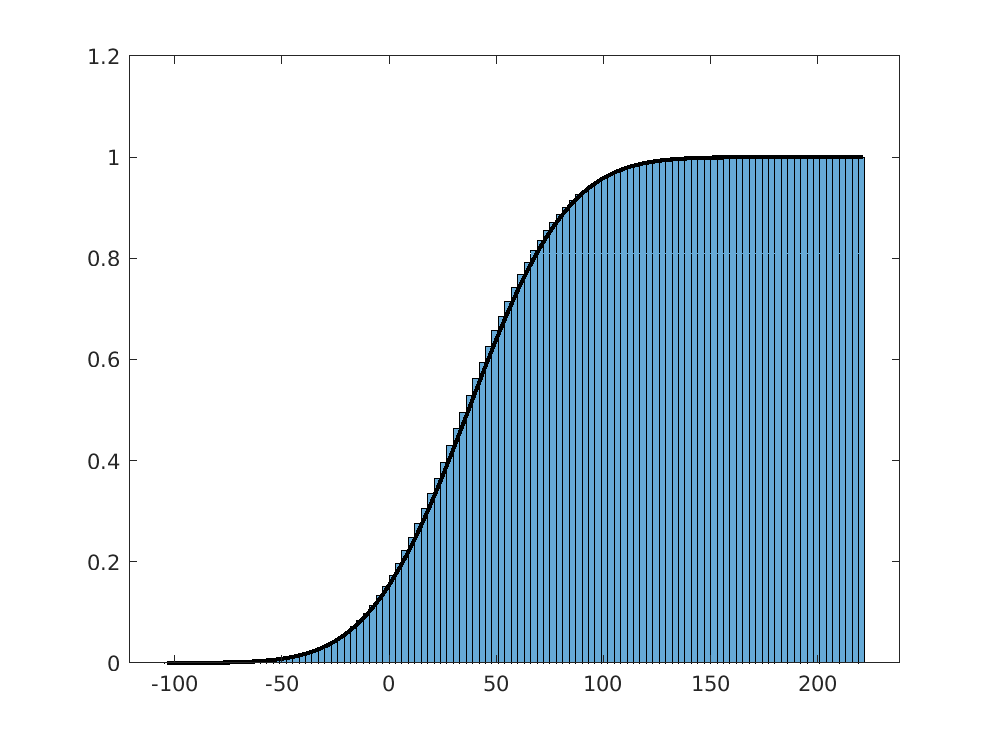}
     \end{subfigure}
     \begin{subfigure}[b]{0.19\linewidth}
         \centering
         \includegraphics[width=\linewidth]{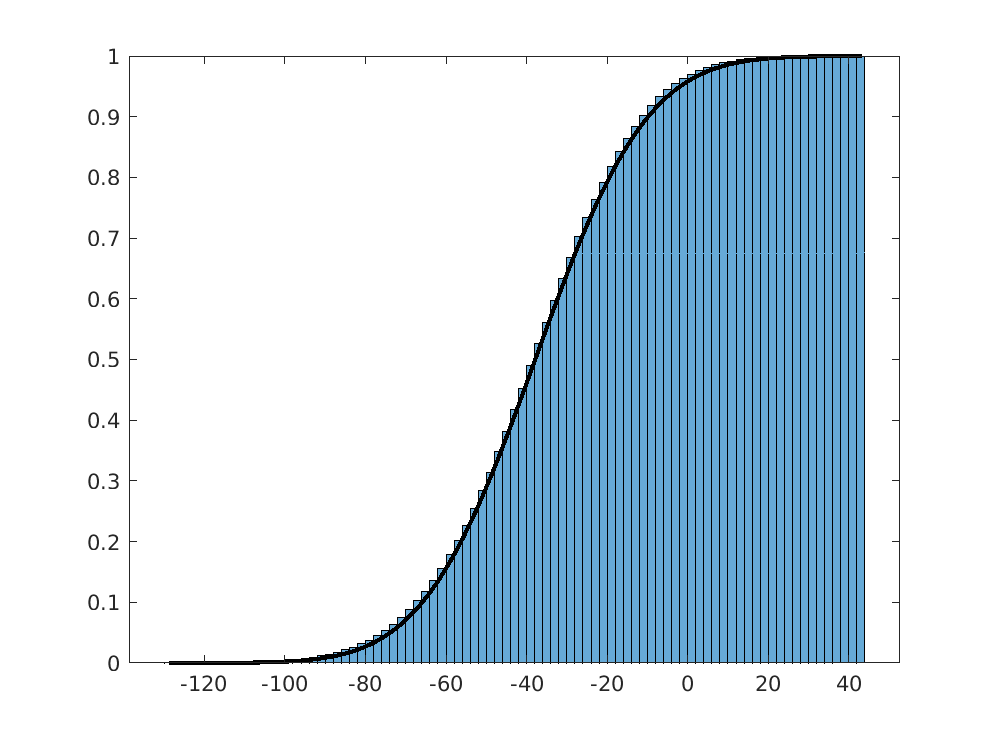}
     \end{subfigure}
     \begin{subfigure}[b]{0.19\linewidth}
         \centering
         \includegraphics[width=\linewidth]{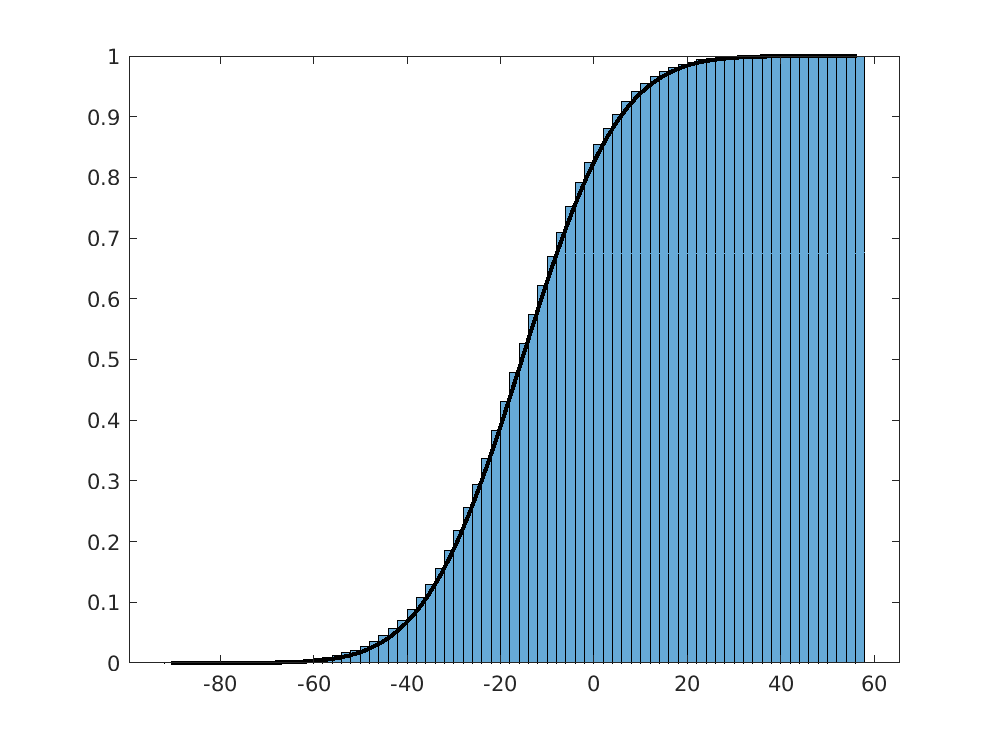}
     \end{subfigure}
     \caption{$D = 32$}
\end{figure}
\begin{figure}[htp!]
    \centering
     \begin{subfigure}[b]{0.19\linewidth}
         \centering
         \includegraphics[width=\linewidth]{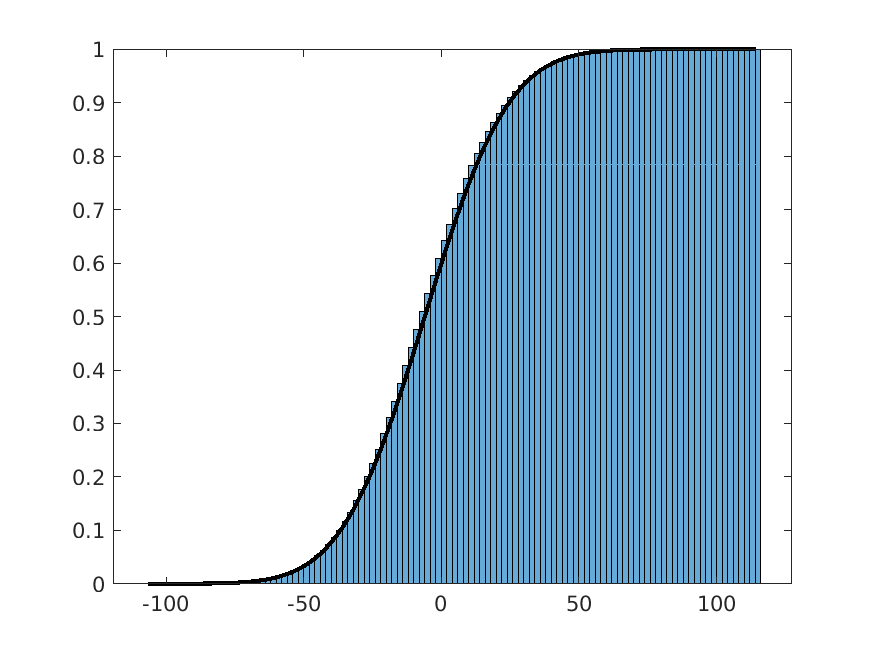} 
     \end{subfigure}
     \begin{subfigure}[b]{0.19\linewidth}
         \centering
         \includegraphics[width=\linewidth]{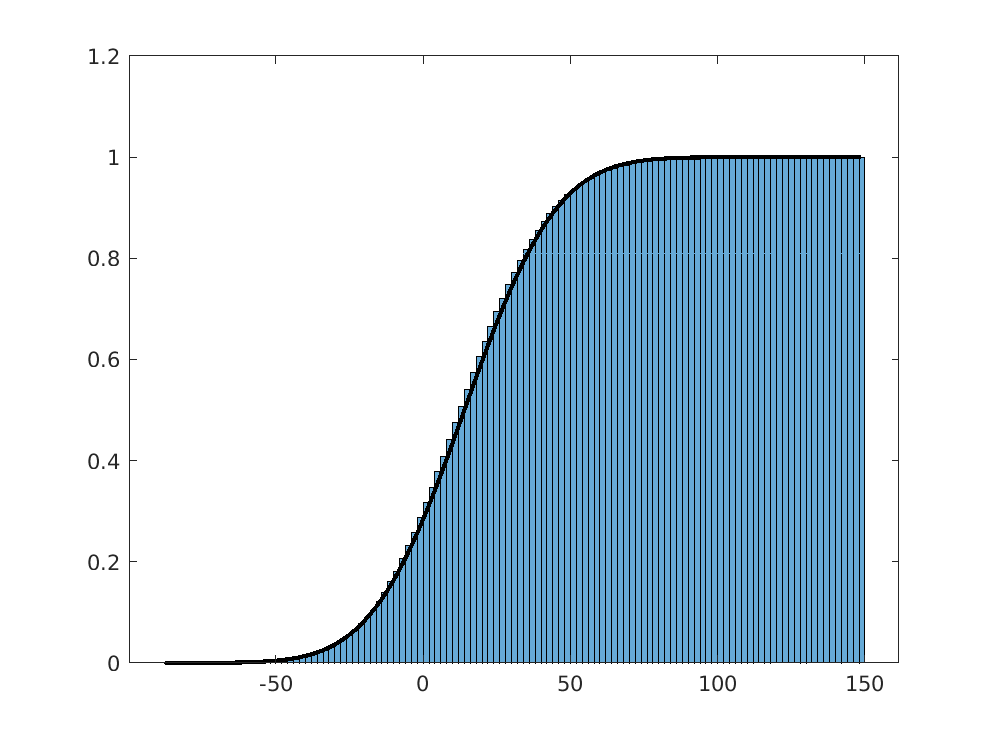}
     \end{subfigure}
     \begin{subfigure}[b]{0.19\linewidth}
         \centering
         \includegraphics[width=\linewidth]{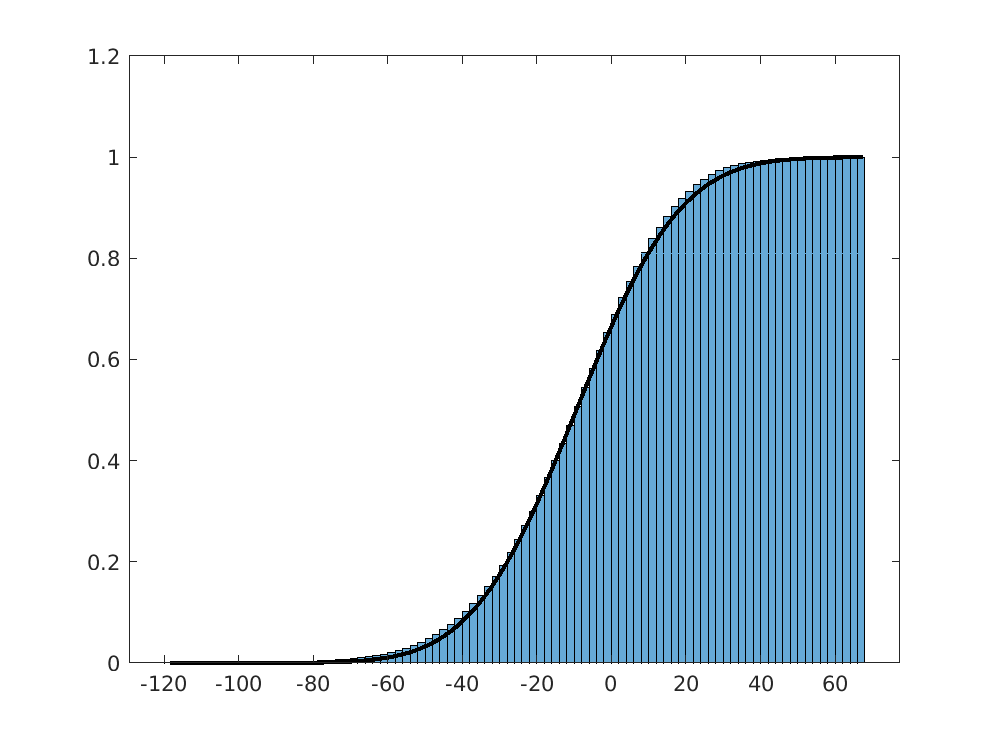}
     \end{subfigure}
     \begin{subfigure}[b]{0.19\linewidth}
         \centering
         \includegraphics[width=\linewidth]{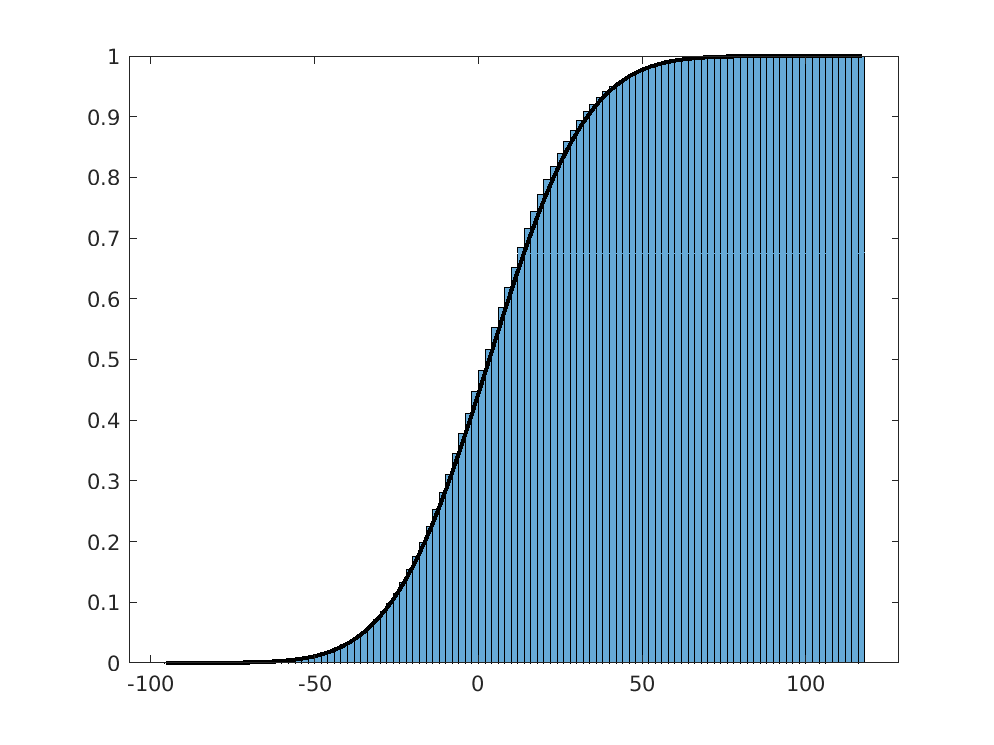}
     \end{subfigure}
     \begin{subfigure}[b]{0.19\linewidth}
         \centering
         \includegraphics[width=\linewidth]{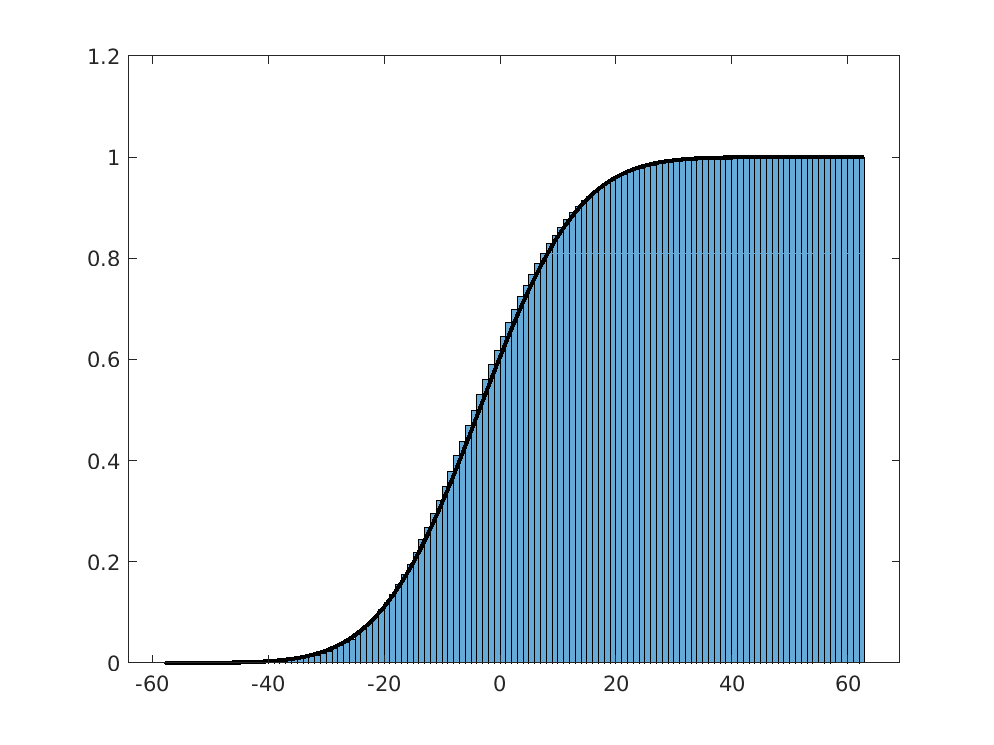}
     \end{subfigure}
     \caption{$D = 16$}     
\end{figure}
\begin{figure}[htp!]
    \centering
     \begin{subfigure}[b]{0.19\linewidth}
         \centering
         \includegraphics[width=\linewidth]{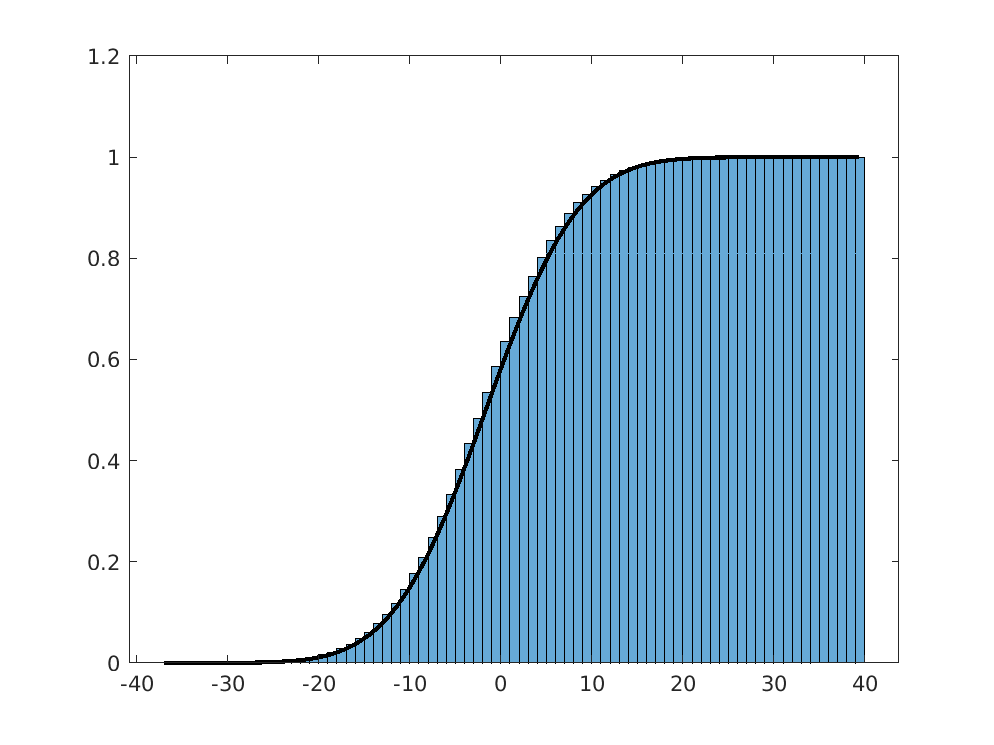} 
     \end{subfigure}
     \begin{subfigure}[b]{0.19\linewidth}
         \centering
         \includegraphics[width=\linewidth]{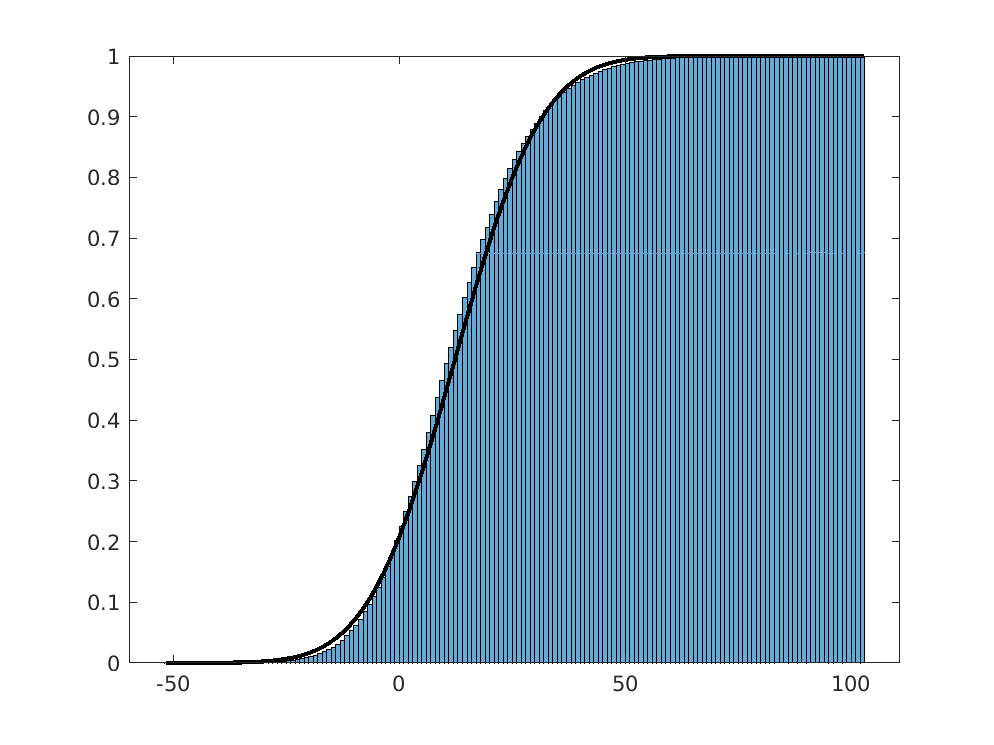}
     \end{subfigure}
     \begin{subfigure}[b]{0.19\linewidth}
         \centering
         \includegraphics[width=\linewidth]{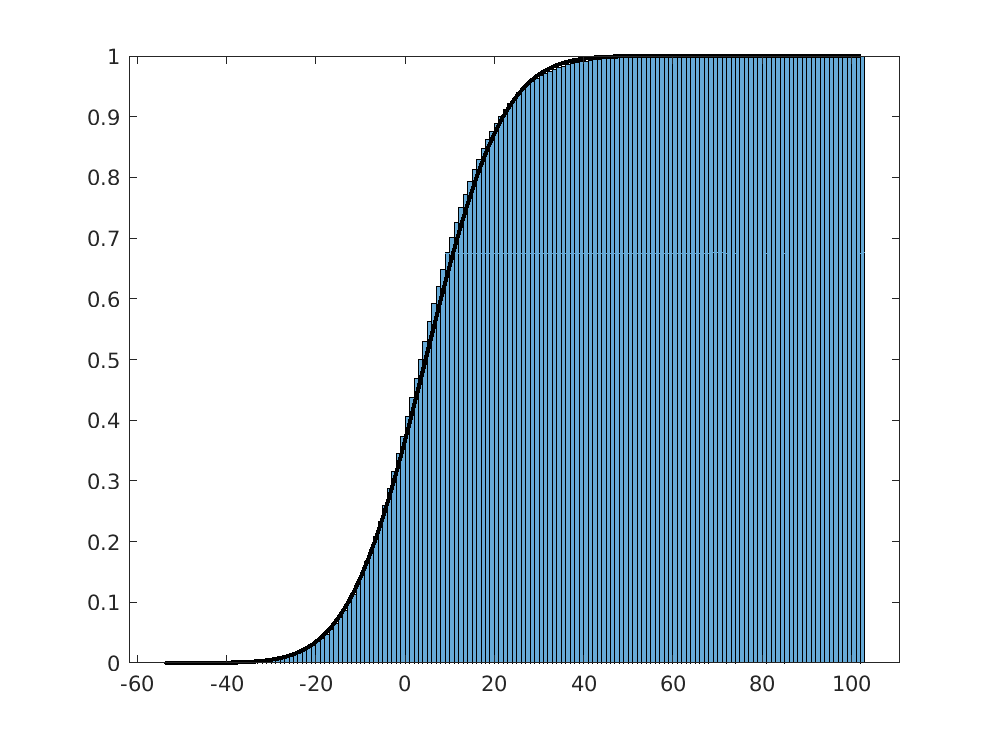}
     \end{subfigure}
     \begin{subfigure}[b]{0.19\linewidth}
         \centering
         \includegraphics[width=\linewidth]{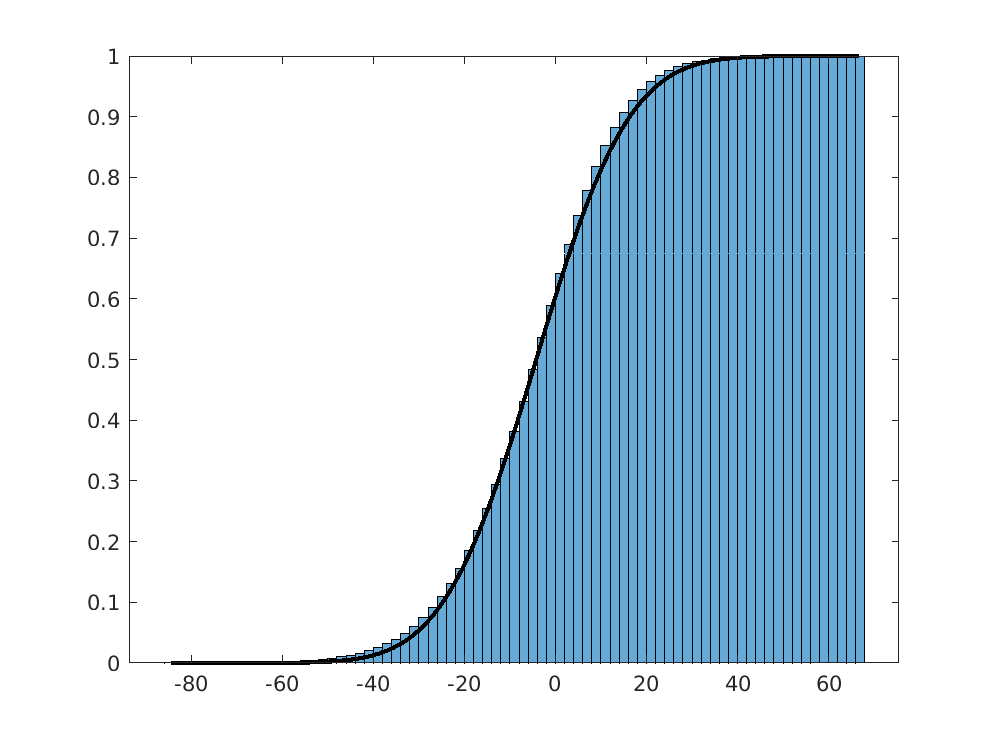}
     \end{subfigure}
     \begin{subfigure}[b]{0.19\linewidth}
         \centering
         \includegraphics[width=\linewidth]{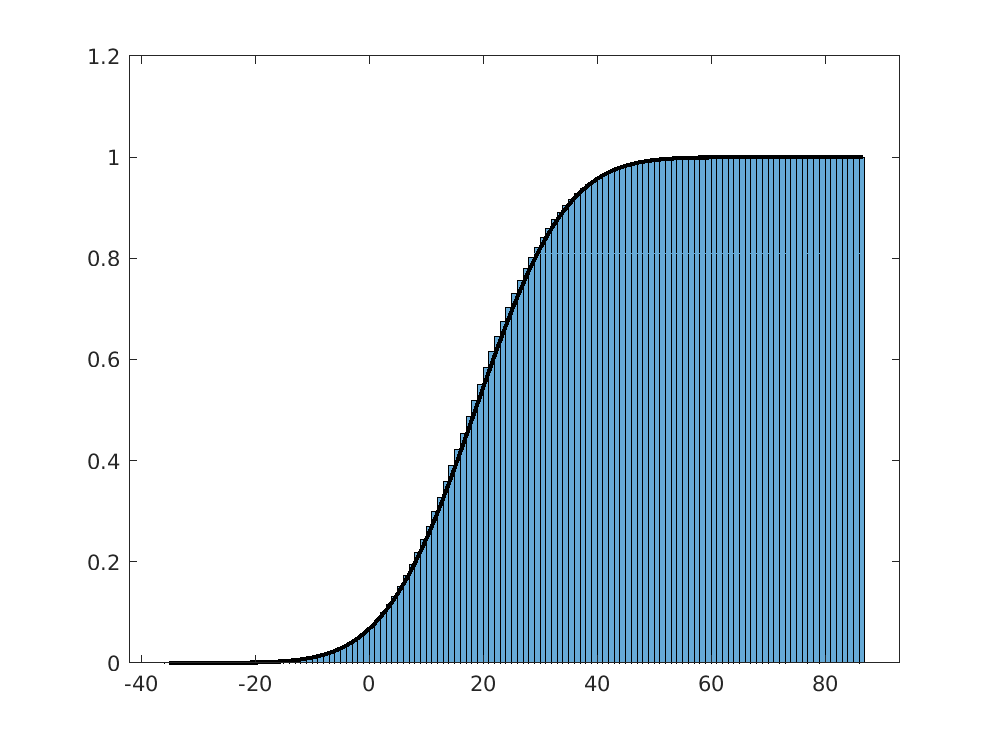}
     \end{subfigure}
     \caption{$D = 8$}
\end{figure}
\begin{figure}[htp!]
    \centering
     \begin{subfigure}[b]{0.19\linewidth}
         \centering
         \includegraphics[width=\linewidth]{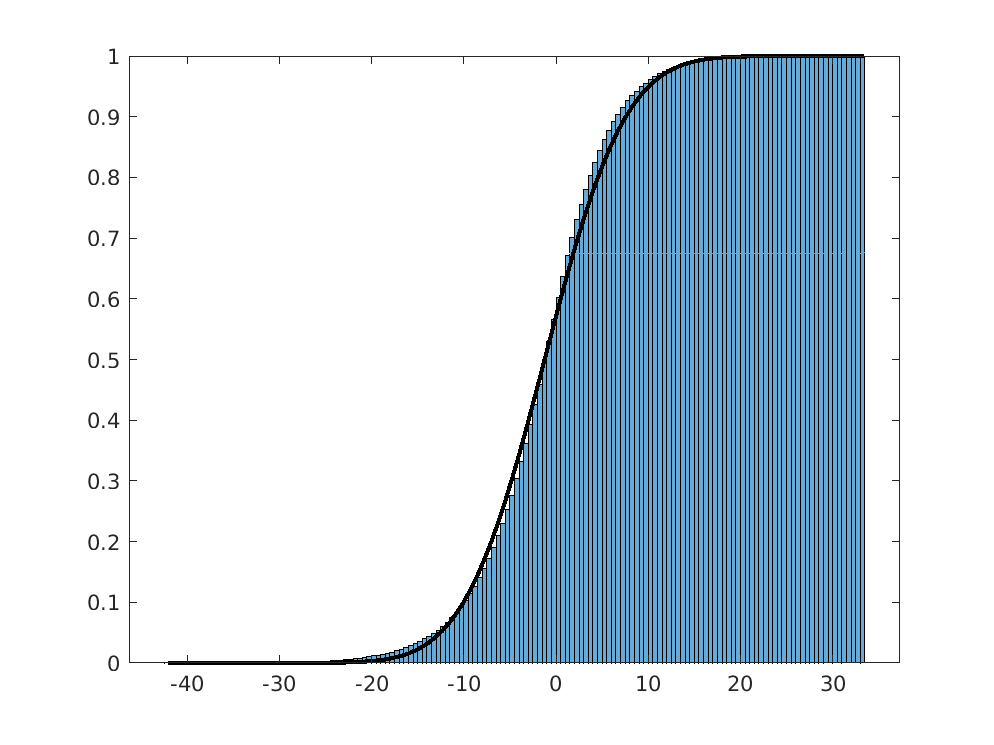} 
     \end{subfigure}
     \begin{subfigure}[b]{0.19\linewidth}
         \centering
         \includegraphics[width=\linewidth]{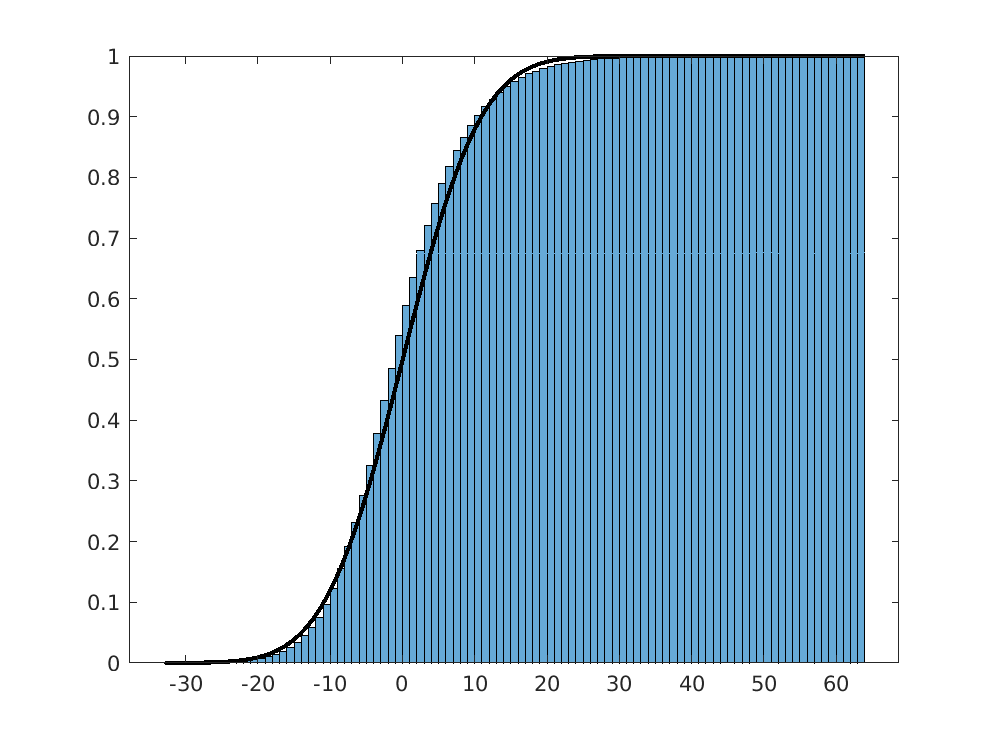}
     \end{subfigure}
     \begin{subfigure}[b]{0.19\linewidth}
         \centering
         \includegraphics[width=\linewidth]{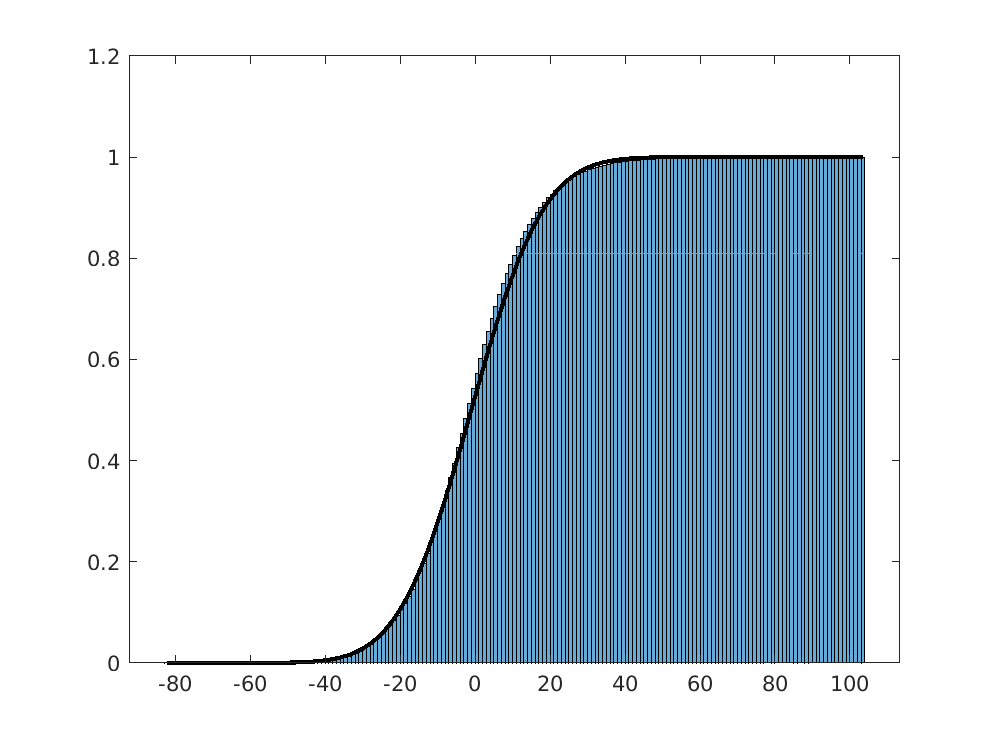}
     \end{subfigure}
     \begin{subfigure}[b]{0.19\linewidth}
         \centering
         \includegraphics[width=\linewidth]{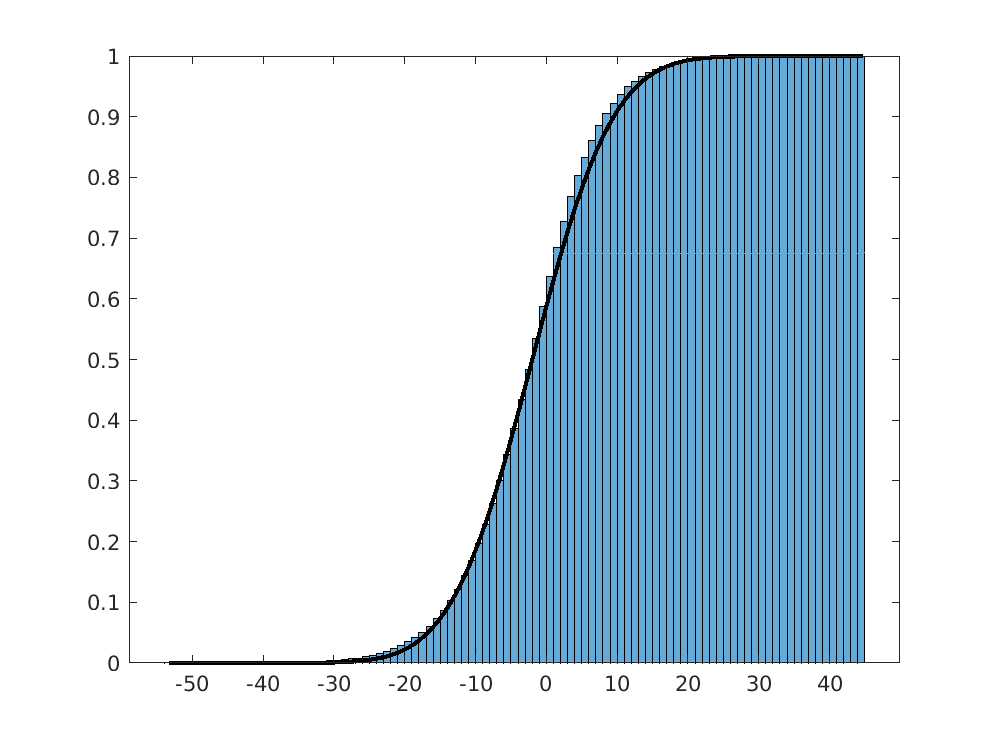}
     \end{subfigure}
     \begin{subfigure}[b]{0.19\linewidth}
         \centering
         \includegraphics[width=\linewidth]{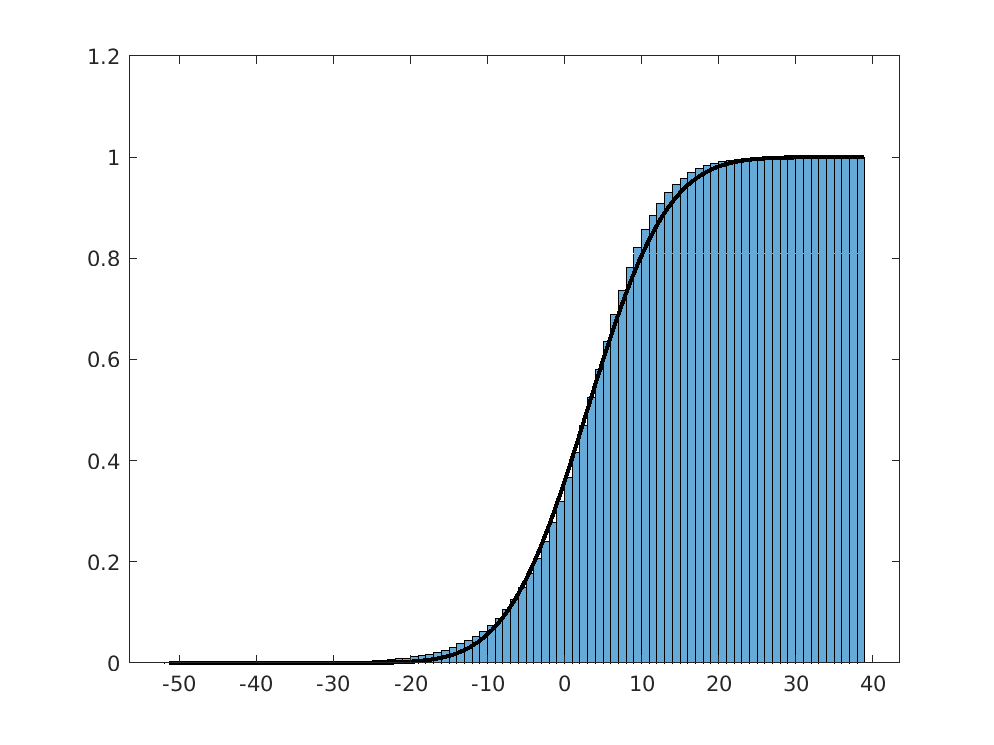}
     \end{subfigure}
     \caption{$D = 4$}
\end{figure}
\begin{figure}[htp!]
    \centering
     \begin{subfigure}[b]{0.19\linewidth}
         \centering
         \includegraphics[width=\linewidth]{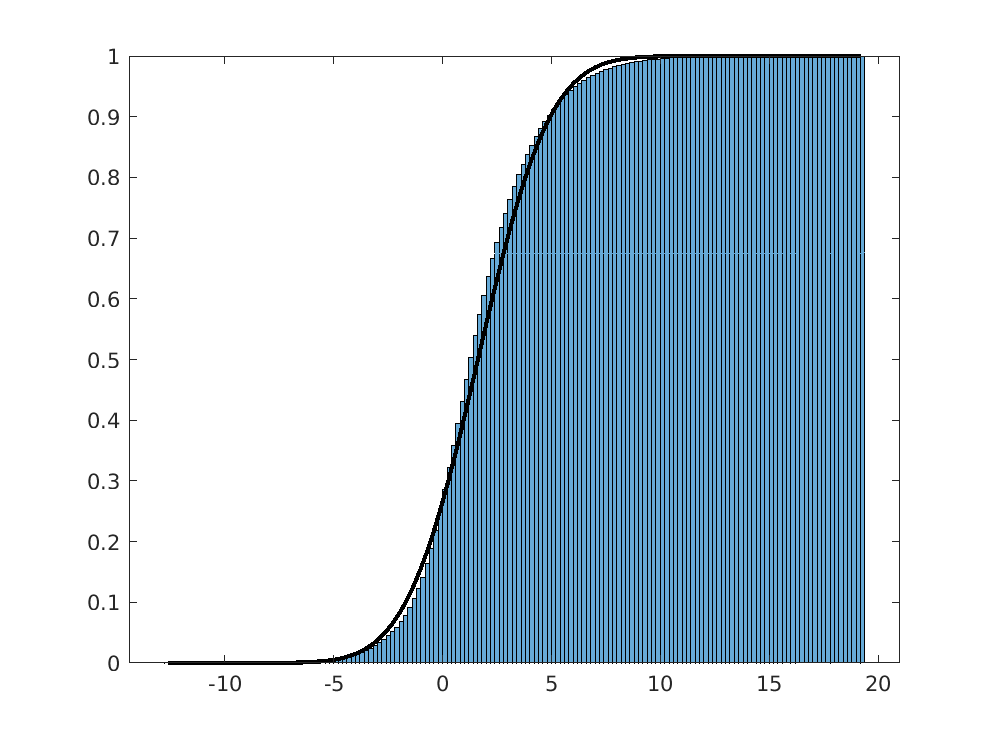} 
     \end{subfigure}
     \begin{subfigure}[b]{0.19\linewidth}
         \centering
         \includegraphics[width=\linewidth]{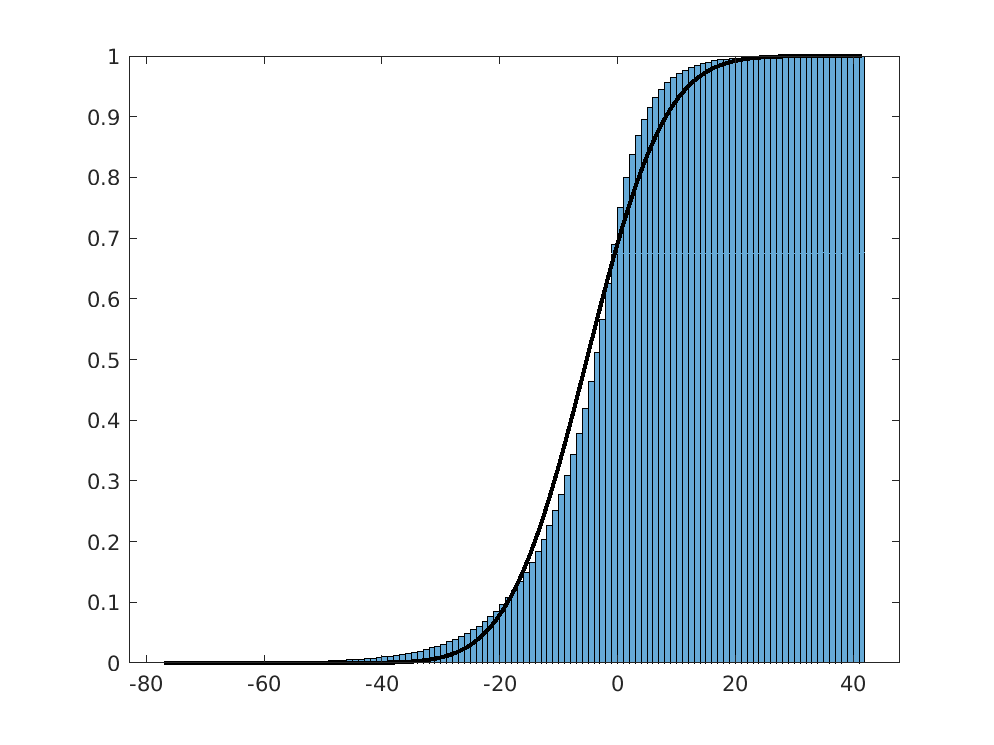}
     \end{subfigure}
     \begin{subfigure}[b]{0.19\linewidth}
         \centering
         \includegraphics[width=\linewidth]{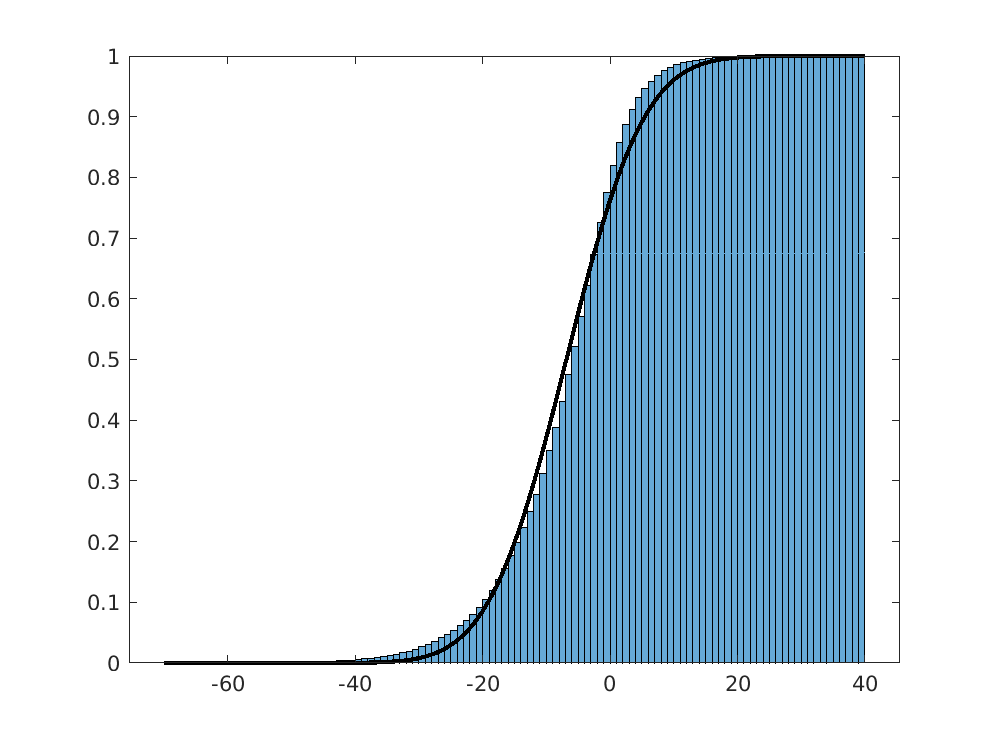}
     \end{subfigure}
     \begin{subfigure}[b]{0.19\linewidth}
         \centering
         \includegraphics[width=\linewidth]{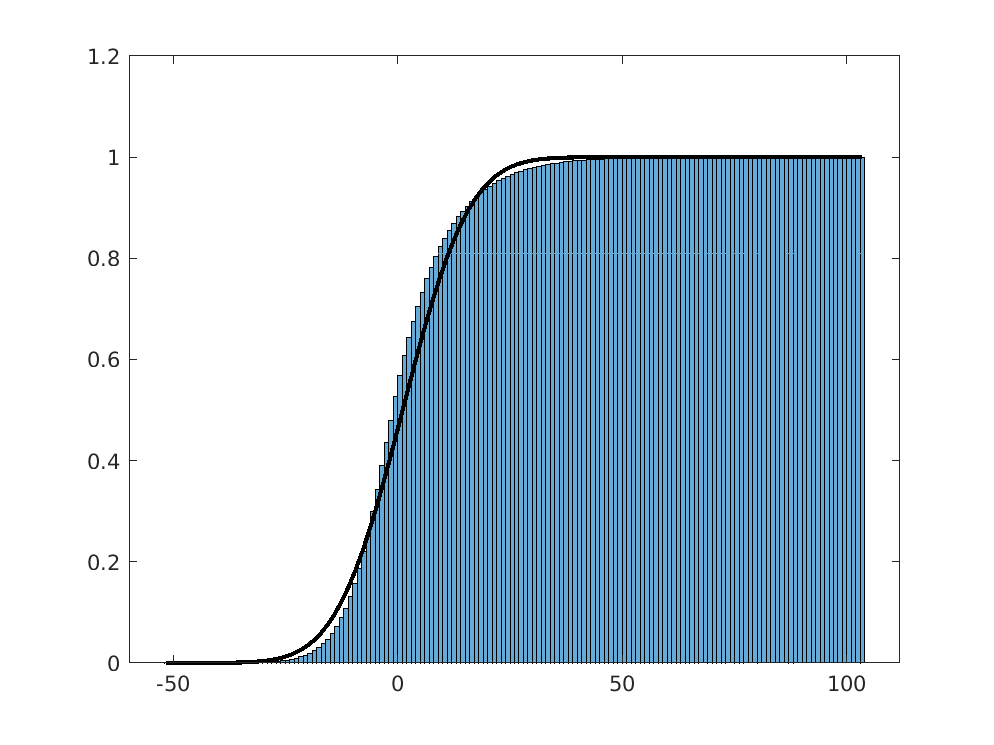}
     \end{subfigure}
     \begin{subfigure}[b]{0.19\linewidth}
         \centering
         \includegraphics[width=\linewidth]{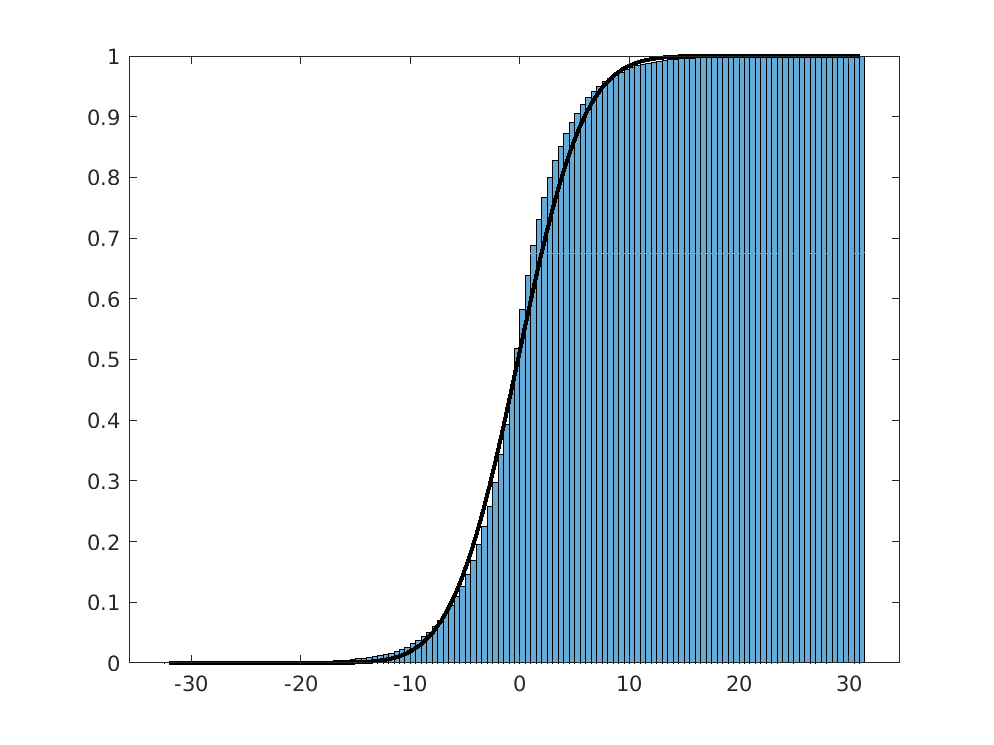}
     \end{subfigure}
     \caption{$D = 2$}
\end{figure}
\begin{figure}[htp!]
    \centering
     \begin{subfigure}[b]{0.19\linewidth}
         \centering
         \includegraphics[width=\linewidth]{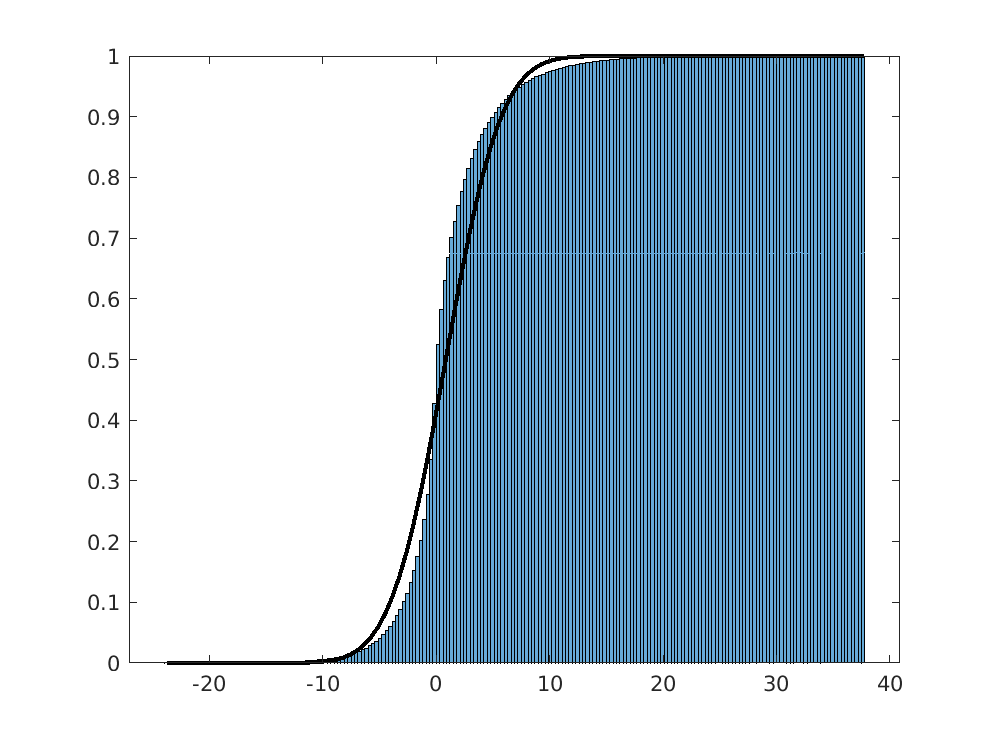} 
     \end{subfigure}
     \begin{subfigure}[b]{0.19\linewidth}
         \centering
         \includegraphics[width=\linewidth]{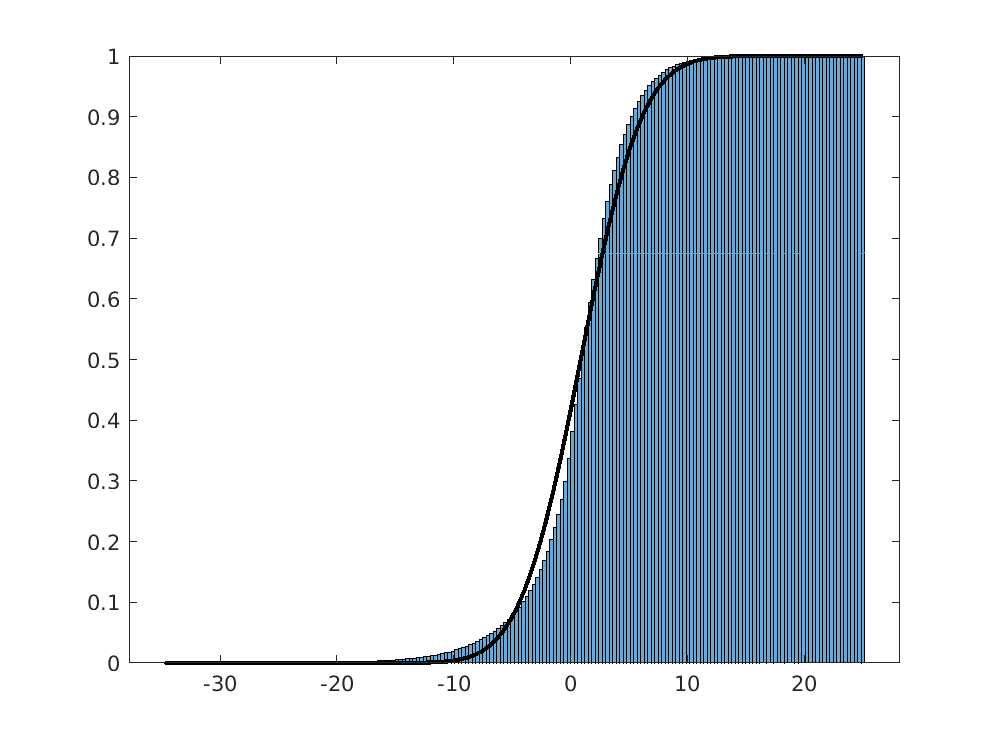}
     \end{subfigure}
     \begin{subfigure}[b]{0.19\linewidth}
         \centering
         \includegraphics[width=\linewidth]{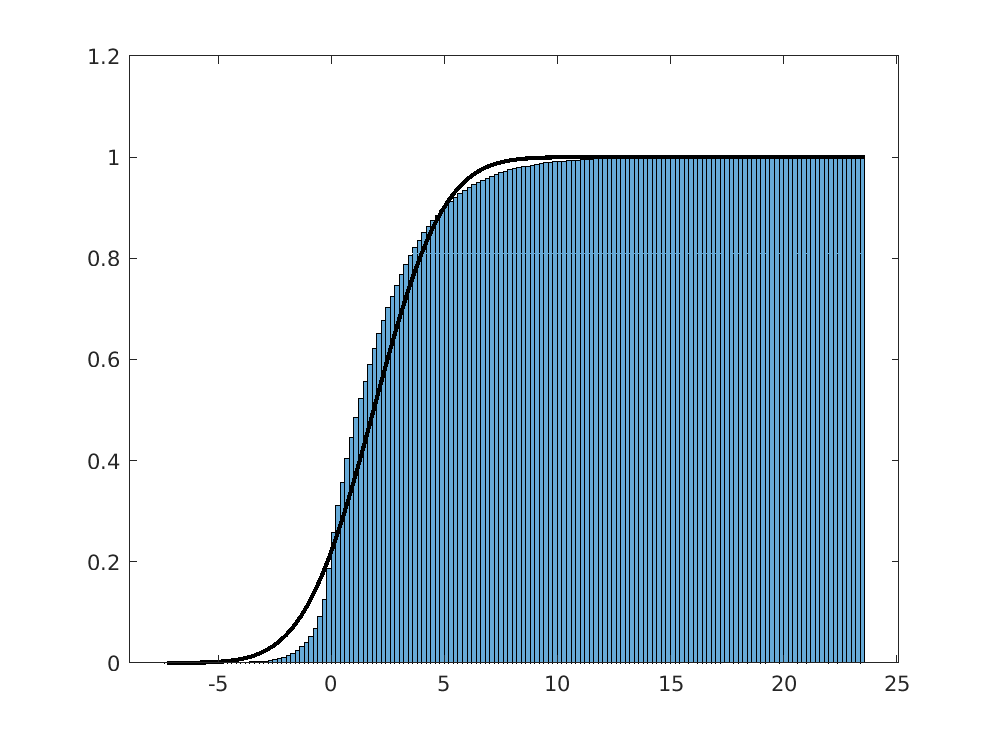}
     \end{subfigure}
     \begin{subfigure}[b]{0.19\linewidth}
         \centering
         \includegraphics[width=\linewidth]{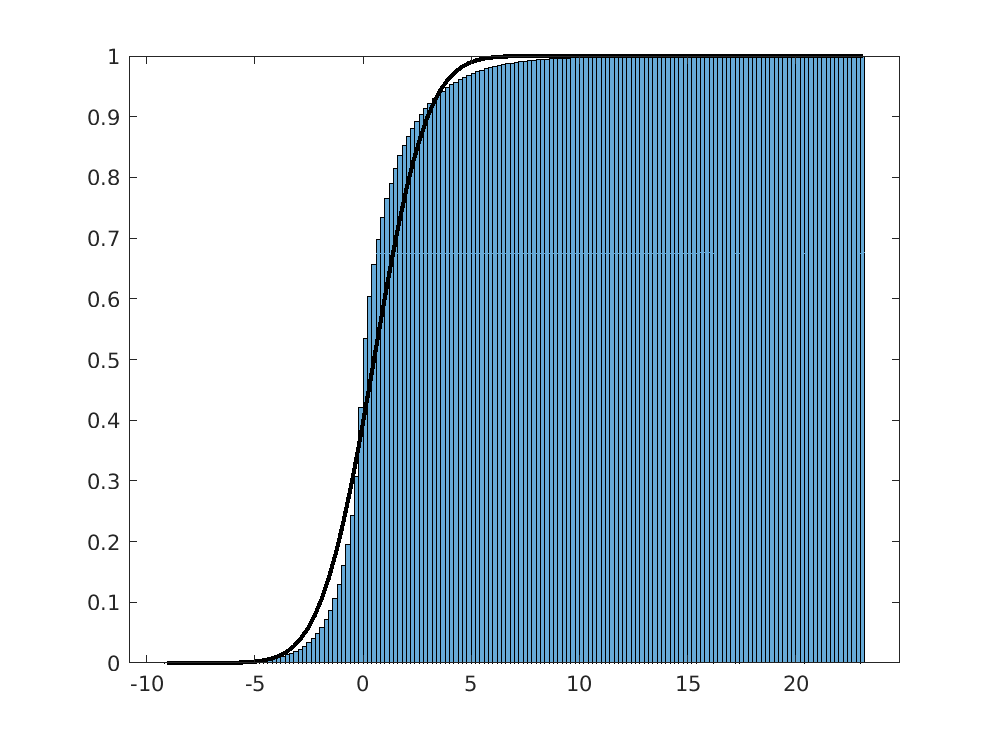}
     \end{subfigure}
     \begin{subfigure}[b]{0.19\linewidth}
         \centering
         \includegraphics[width=\linewidth]{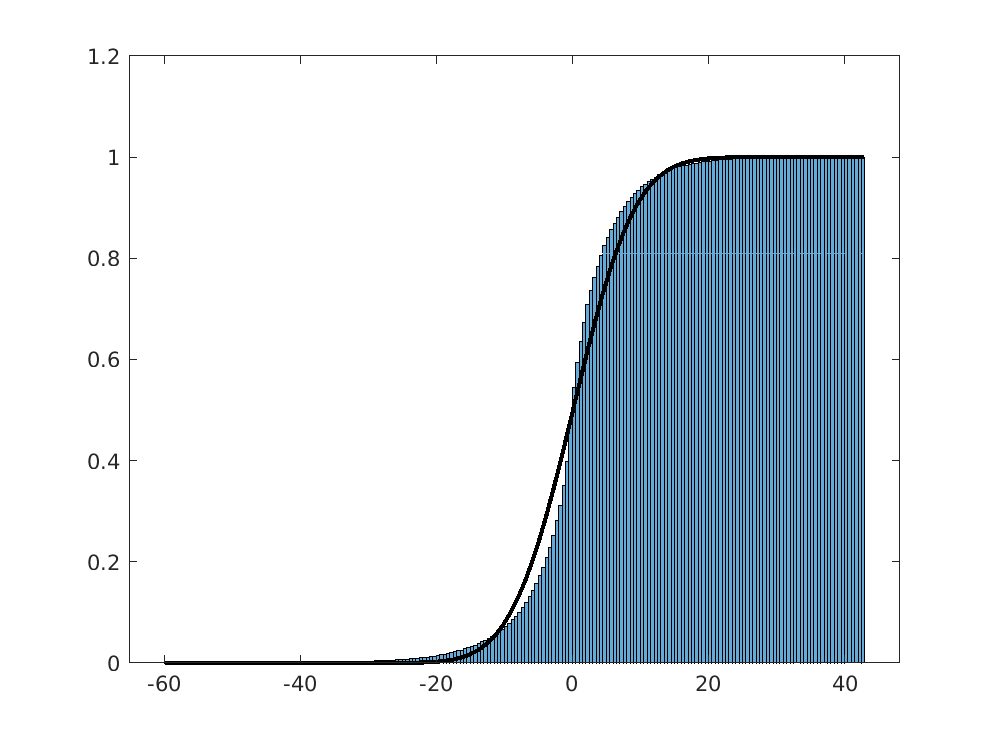}
     \end{subfigure}
     \caption{$D = 1$}
\end{figure}

\section{Variational Approximation}
In this section we derive the Expected Lower Bound. We start with the marginal probability of a triplet label $I(X, Y, Z)$.

\begin{align}
    log P( I(X, Y, Z) ) & = log \int P( I(X, Y, Z), (X, Y, Z)) d(X, Y, Z) \\
                        & = log \int \int \int P( I(X, Y, Z), X, Y, Z) dX dY dZ \\
                        & = log \int \int \int P( I(X, Y, Z) | X, Y, Z) P(X, Y, Z) dX dY dZ \\
                        & = log \int \int \int P( I(X, Y, Z) | X, Y, Z) P(X) P(Y) P(Z) dX dY dZ \\
                        & = log \int \int \int \frac{P( I(X, Y, Z) | X, Y, Z) P(X) P(Y) P(Z) }{q(X) q(Y) q(Z)} q(X) q(Y) q(Z) dX dY dZ \\
                        & \geq \int \int \int log \frac{P( I(X, Y, Z) | X, Y, Z) P(X) P(Y) P(Z) }{q(X) q(Y) q(Z)} q(X) q(Y) q(Z) dX dY dZ \\
                        & = \int \int \int log P( I(X, Y, Z) | X, Y, Z) q(X) q(Y) q(Z) \\
                        & + log \frac{P(X) P(Y) P(Z)}{q(X) q(Y) q(Z)} q(X) q(Y) q(Z)dX dY dZ \\
                        & = \int \int \int log P( I(X, Y, Z) | X, Y, Z) q(X) q(Y) q(Z) dX dY dZ \\
                        & + \int \int \int log \frac{P(X) P(Y) P(Z)}{q(X) q(Y) q(Z)} q(X) q(Y) q(Z) dX dY dZ \\
\end{align}

We have two terms. The first part becomes

\begin{align}
       & \int \int \int log P( I(X, Y, Z) | X, Y, Z) q(X) q(Y) q(Z) dX dY dZ \\
    =  & \mathbb{E}_{q(X)q(Y)q(Z)}[log P(I(X, Y, Z) | X, Y, Z) ] 
\end{align}

The second part becomes

\begin{align}
      & \int \int \int log \frac{P(X) P(Y) P(Z)}{q(X) q(Y) q(Z)}q(X) q(Y) q(Z) dX dY dZ \\
    = &\int \int \int [ log P(X) - log q(X) + log P(Y) - log q(Y) + log P(Z) - log q(Z) ] q(X) q(Y) q(Z) dX dY dZ \\
    = & \int \int \int [log P(X) - log q(X)] q(X) q(Y) q(Z) dX dY dZ \\
      & + \int \int \int [log P(Y) - log q(Y)]q(X) q(Y) q(Z) dX dY dZ \\ 
      & + \int \int \int  [log P(Z) - log q(Z)] q(X) q(Y) q(Z) dX dY dZ \\
    = & \int [log P(X) - log q(X)] q(X) dX + \int [log P(Y) - log q(Y)] q(Y) dY  + \int [log P(Z) - log q(Z)] q(Z) dZ \\
    = & - KL(q(X) || P(X)) - KL(q(Y) || P(Y)) - KL(q(Z) || P(Z))
 \end{align}

This means that if we choose $P(X) = P(Y) = P(Z) = N(0, 1)$, we can remove the normalization layer and optimize the ELBO directly.

\section{Implementation}
We present our Pytorch implementation of the Bayesian triplet loss with both the von Mises Fischer and Gaussian embeddings. We will make the source code available upon acceptance.

\begin{lstlisting}[language=Python, label=loss,caption=loss.py]
import torch
import torch.nn as nn
import functional as LF

class BayesianTripletLoss(nn.Module):

    def __init__(self, margin, varPrior, kl_scale_factor = 1e-6, distribution = 'gauss'):
        super(BayesianTripletLoss, self).__init__()

        self.margin = margin
        self.varPrior = varPrior
        self.kl_scale_factor = kl_scale_factor
        self.distribution = distribution

    def forward(self, x, label):
        
        # divide x into anchor, positive, negative based on labels
        D, N = x.shape
        nq = torch.sum(label.data == -1).item()  # number of tuples
        S = x.size(1) // nq  # number of images per tuple including query: 1+1+n
        A = x[:, label.data == -1].permute(1, 0).repeat(1, S - 2).view((S - 2) * nq, D).permute(1, 0)
        P = x[:, label.data == 1].permute(1, 0).repeat(1, S - 2).view((S - 2) * nq, D).permute(1, 0)
        N = x[:, label.data == 0]

        varA = A[-1:, :]
        varP = P[-1:, :]
        varN = N[-1:, :]

        muA = A[:-1, :]
        muP = P[:-1, :]
        muN = N[:-1, :]

        # calculate nll
        nll = LF.negative_loglikelihood(muA, muP, muN, varA, varP, varN, margin=self.margin)
        
        # KL(anchor|| prior) + KL(positive|| prior) + KL(negative|| prior)        
        if self.distribution == 'gauss':

            muPrior = torch.zeros_like(muA, requires_grad = False)
            varPrior = torch.ones_like(varA, requires_grad = False) * self.varPrior

            kl = (LF.kl_div_gauss(muA, varA, muPrior, varPrior) + \
                    LF.kl_div_gauss(muP, varP, muPrior, varPrior) + \
                    LF.kl_div_gauss(muN, varN, muPrior, varPrior))

        elif self.distribution == 'vMF':
            kl = (LF.kl_div_vMF(muA, varA) + \
                    LF.kl_div_vMF(muP, varP) + \
                    LF.kl_div_vMF(muN, varN))
        
        return nll + self.kl_scale_factor * kl 

    def __repr__(self):
        return self.__class__._Name__ + '(' + 'margin=' + '{:.4f}'.format(self.margin) + ')'
\end{lstlisting}
\begin{lstlisting}[language=Python, label=functional,caption=functional.py]
import torch
from torch.distributions.normal import Normal

def negative_loglikelihood(muA, muP, muN, varA, varP, varN, margin = 0.0):

    muA2 = muA**2
    muP2 = muP**2
    muN2 = muN**2
    varP2 = varP**2
    varN2 = varN**2
    
    mu = torch.sum(muP2 + varP - muN2 - varN - 2*muA*(muP - muN), dim=0)
    T1 = varP2 + 2*muP2 * varP + 2*(varA + muA2)*(varP + muP2) - 2*muA2 * muP2 - 4*muA*muP*varP
    T2 = varN2 + 2*muN2 * varN + 2*(varA + muA2)*(varN + muN2) - 2*muA2 * muN2 - 4*muA*muN*varN
    T3 = 4*muP*muN*varA
    sigma2 = torch.sum(2*T1 + 2*T2 - 2*T3, dim=0)
    sigma = sigma2**0.5

    probs = Normal(loc = mu, scale = sigma + 1e-8).cdf(margin)    
    nll = -torch.log(probs + 1e-8)

    return nll.mean()

def kl_div_gauss(mu_q, var_q, mu_p, var_p):

    N, D = mu_q.shape

    # kl diverence for isotropic gaussian
    kl = 0.5 * ((var_q / var_p) * D + \
                1.0 / (var_p) * torch.sum(mu_p**2 + mu_q**2 - 2*mu_p*mu_q, axis=1) - D + \
                D*(torch.log(var_p) - torch.log(var_q)))

    return kl.mean()

def kl_div_vMF(mu_q, var_q):
    
    N, D = mu_q.shape

    # we are estimating the variance and not kappa in the network. 
    # They are propertional
    kappa_q = 1.0 / var_q
    kl = kappa_q - D * torch.log(2.0)
    
    return kl.mean()
\end{lstlisting}
